\documentclass[mnsc,nonblindrev]{informs3_hide} %

\OneAndAHalfSpacedXI %

\usepackage{amsmath,amsfonts,amssymb}
\usepackage{mathtools}
\allowdisplaybreaks
\usepackage{bm}
\usepackage[mathscr]{euscript}
\usepackage{enumitem}
\usepackage{xcolor}
\usepackage{graphicx}
\usepackage{multirow}
\usepackage{booktabs}
\usepackage{microtype}
\usepackage{fix-cm}
\usepackage{physics}
\usepackage{xfrac}
\usepackage{comment}
\usepackage{extarrows}

\usepackage{algorithm}
\usepackage{algpseudocode}
\usepackage{caption, subcaption}

\newcommand{\bx}{\mathbf{x}}

\newcommand{\by}{\mathbf{y}}
\newcommand{\bY}{\mathbf{Y}}

\newcommand{\bb}{\mathbf{b}}

\newcommand{\bd}{\mathbf{d}}
\newcommand{\bB}{\mathbf{B}}
\newcommand{\bxi}{\boldsymbol{\xi}}
\newcommand{\balpha}{\boldsymbol{\alpha}}
\newcommand{\bbeta}{\boldsymbol{\beta}}

\newcommand{\bgamma}{\boldsymbol{\gamma}}

\makeatletter
\newcommand{\biggg}{\bBigg@{3}}

\newcommand{\Biggg}{\bBigg@{3.5}}

\newcommand{\bigggg}{\bBigg@{4}}

\newcommand{\Bigggg}{\bBigg@{4.5}}

\makeatother

\usepackage{hyperref}
\hypersetup{hidelinks,
	colorlinks=true,
	allcolors=black,
	pdfstartview=Fit,
	breaklinks=true}

\usepackage{natbib}
\bibpunct[, ]{(}{)}{,}{a}{}{,}%
\def\bibfont{\small}%
\def\bibsep{\smallskipamount}%
\TheoremsNumberedThrough     %
\ECRepeatTheorems

\EquationsNumberedThrough    %

\begin{document}

\RUNAUTHOR{Hong et al.}
	
\RUNTITLE{Learning to Simulate via Quantile Regression}
	
\TITLE{Learning to Simulate: Generative Metamodeling via Quantile Regression}
	
\ARTICLEAUTHORS{
	\AUTHOR{L. Jeff Hong}
	\AFF{Department of Industrial and Systems Engineering, University of Minnesota, Minneapolis, MN 55455, U.S., \EMAIL{lhong@umn.edu}}
	\AUTHOR{Yanxi Hou}
	\AFF{School of Data Science, Fudan University, Shanghai 200433, China, \EMAIL{yxhou@fudan.edu.cn}}
	\AUTHOR{Qingkai Zhang}
	\AFF{School of Management, Fudan University, Shanghai 200433, China, \EMAIL{22110690021@m.fudan.edu.cn}}
	\AFF{Department of Decision Analytics and Operations, City University of Hong Kong,
		Kowloon, Hong Kong SAR, China}
	\AUTHOR{Xiaowei Zhang}
	\AFF{Department of Industrial Engineering and Decision Analytics, The Hong Kong University of Science and Technology, Clear Water Bay, Hong Kong SAR, China, \EMAIL{xiaoweiz@ust.hk}}
}
	
\ABSTRACT{Stochastic simulation models effectively capture complex system dynamics but are often too slow for real-time decision-making. Traditional metamodeling techniques learn relationships between simulator inputs and a single output summary statistic, such as the mean or median. These techniques enable real-time predictions without additional simulations. However, they require prior selection of one appropriate output summary statistic, limiting their flexibility in practical applications. We propose a new concept: \emph{generative metamodeling}. It aims to construct a ``fast simulator of the simulator," generating random outputs significantly faster than the original simulator while preserving approximately equal conditional distributions. Generative metamodels enable rapid generation of numerous random outputs upon input specification, facilitating immediate computation of any summary statistic for real-time decision-making. We introduce a new algorithm, quantile-regression-based generative metamodeling (QRGMM), and establish its distributional convergence. Extensive numerical experiments demonstrate QRGMM's efficacy compared to other state-of-the-art generative algorithms in practical real-time decision-making scenarios.
}
	

\maketitle

\section{Introduction}\label{sec:intro}

Simulation is a powerful tool for modeling and analyzing real-world systems. However, running complex simulation models is often time-consuming, limiting its use to design-phase applications where sufficient time exists for system evaluation, improvement, or optimization. In recent years, the demand for simulation in real-time decision-making has grown, particularly in areas like digital twins in manufacturing and real-time antenna control in telecommunications. The challenge is to bridge the gap between computationally intensive simulations and the need for rapid, on-the-fly decisions under tight time constraints.

In such scenarios, a subset of simulation input variables, termed \emph{covariates}, are observed in real time. These covariates provide critical contextual information for decision-making while introducing temporal constraints, as their values are unknown until observation, requiring swift responses. To address this, \cite{hong2019offline} proposed an offline-simulation-online-application (OSOA) framework. In the offline stage, extensive simulation experiments are conducted over a wide range of covariate values to build a \emph{metamodel} (predictive model) that captures the relationship between covariates and a pre-specified summary statistic of simulation outputs. In the online stage, the metamodel predicts the summary statistic rapidly upon observing new covariate values, bypassing the need to run the simulator, thus enabling real-time decisions. For recent studies applying similar approaches to address decision-making problems in inventory management, financial engineering, and health care, we refer to \cite{HannahPowellDunson14,jiang2020online}, and \cite{shen2021ranking}.

Metamodeling is central to the OSOA framework, with techniques like linear regression, stochastic kriging, and neural networks commonly used \citep{barton2020tutorial}.
However, a significant limitation of traditional metamodeling techniques is the necessity to specify the summary statistic of simulation outputs in advance, before collecting data to build the metamodel. For example, \cite{jiang2020online} focused on the probability of financial portfolio losses exceeding a threshold, and \cite{shen2021ranking} used the expected quality-adjusted life years (QALYs) for patients under specific treatment regimes.

This approach may be unsuitable for many practical real-time decision-making problems where varying scenarios necessitate different summary statistics.
In personalized medicine, for instance,
patients have diverse priorities. 
While some focus on expected QALYs, 
others may prioritize the probability of survival beyond a specific critical time point. 
Critically, these individual preferences emerge only during the doctor-patient consultation, that is, after the covariate values (such as patient characteristics) have been observed.

A fundamental question arises: \emph{What constitutes an ideal metamodel?} 
To address this, let us consider a hypothetical scenario involving a super-fast computer capable of generating simulation observations instantaneously. 
In such a scenario, constructing a metamodel would be unnecessary. 
One could simply wait until the covariate values are observed, then run the simulator to generate the required number of output observations. 
Any desired summary statistics could then be calculated directly from these simulation outputs.
While this scenario is purely hypothetical, it sheds light on the key characteristics of an ideal metamodel.

Let $\bx$ denote the covariates and $Y(\bx)$ denote the \emph{random} output of the simulator. 
An ideal metamodel should be capable of generating random outputs that share the same distribution as $Y(\bx)$ for any given $\bx$, but at a much faster speed.
We may formulate the construction of such a metamodel as using offline simulation data to learn a fast, random function $\hat Y(\bx)$, which approximates $Y(\bx)$ in distribution.
In contrast, a traditional metamodel would approximate the \emph{deterministic} function $\mathcal{S}(Y(\bx))$, where $\mathcal{S}$ represents a summary statistic (e.g., the mean or median) that transforms the distribution of $Y(\bx)$ to a deterministic scalar. 

In essence, our goal is to build a ``fast simulator of the simulator"---a metamodel that is \emph{generative} rather than merely \emph{predictive}. 
With such a generative metamodel, once the covariate values are known, we can quickly generate a large number of (approximate) samples of $Y(\bx)$. 
These samples can then be used to compute any desired summary statistics in real time. 

Our emphasis on modeling the full output distribution, rather than a pre-specified summary statistic, is also aligned with recent work in the simulation community that highlights the limitations of summary-based analysis; see, for example, \citet{chen2022distributional}, who study distributional effects arising from input uncertainty. While their focus is on distributional analysis, our work is complementary in that we aim to construct a fast generative metamodel that can directly generate approximate samples from the conditional output distribution for real-time decision-making.

Constructing a generative metamodel involves learning the conditional distribution of $Y(\bx)$ given $\bx$, a task akin to generative modeling in machine learning. {Techniques like generative adversarial networks (GANs) \citep{goodfellow2014generative}, variational autoencoders \citep{kingma2013auto}, diffusion models \citep{ho2020denoising}, and flow matching models \citep{lipman2023flow} have shown success, particularly in image generation. However, simulation metamodeling has distinct characteristics: (1) \emph{Distributional Proximity}: We prioritize generating a large number of samples that accurately approximate the true distribution of $Y(\bx)$, including its tails. In contrast, generative models in machine learning often aim to produce a few high-quality samples, such as realistic-looking images. Notably, many machine learning methods struggle with mode collapse problems \citep{saxena2021generative}, resulting in generated data that lack the diversity observed in real data and fail to replicate the true distribution. %
(2) \emph{Covariate Nature}: In our applications, covariates are typically continuous variables. Machine learning, however, often deals more with categorical labels. 
While studies utilizing continuous labels have emerged, such as continuous conditional GANs with generator regularization \citep{zheng2021continuous, zhang2022mind}, this setting has received comparatively less attention and warrants further investigation. (3) \emph{Output Dimensionality}: Our focus is on cases where the simulator's output is one-dimensional or low-dimensional, corresponding to most practical simulation problems. 
In contrast, machine learning typically concentrates on higher-dimensional data, such as images, speech, and videos. That said, recent work has started to apply modern generative models to regression problems for learning conditional response distributions and constructing predictive intervals, such as the diffusion-based approach in \citet{han2022card}. Motivated by these advances, we include several representative generative models as benchmarks in the numerical experiments in Section~\ref{sec:num}.

Focusing on one-dimensional outputs (with extensions to multidimensional cases in Section~\ref{sec:extension}), our first main contribution is proposing a \emph{quantile-regression-based generative metamodeling} (QRGMM) approach. Extending the \emph{inverse transform method}, where a random variable $Y$ has the same distribution as $F_Y^{-1}(U)$ with $U \sim \mathsf{Unif}(0,1)$, we learn the conditional quantile function $F_Y^{-1}(\tau|\bx)$ for $\tau\in(0,1)$ via quantile regression on an offline dataset $\{(\bx_i, y_i)\}_{i=1}^{n}$. The resulting estimate $\hat{Q}(\tau|\bx)$ allows generating samples $\hat{Y}(\bx) = \hat{Q}(U|\bx)$ in real time. A practical challenge arises: we must efficiently compute $\hat Q(U|\bx)$ for arbitrary $U$ during real-time operation. Our solution is simple yet effective: offline, we fit quantile regressions at a grid of levels ${\tau_j = j/m}$ for $j=1,\ldots,m-1$; online, we interpolate these predictions to approximate $\hat Q(U|\bx)$.

Our second contribution is the theoretical analysis of QRGMM. Under the premise that $F^{-1}_Y(\tau|\bx)$ is correctly specified by a linear quantile regression model for all $\tau\in(0,1)$, we prove that the distribution of $\hat Y(\bx)$ converges uniformly to that of $Y(\bx)$ as both the simulation sample size $n$ and the number of quantile levels $m$ grow. A key technical challenge arises because prior studies on quantile regression typically assume a fixed quantile level $\tau$ or, in the case of varying quantile levels, assume $\tau$ lies in an interval $[\tau_{\mathsf{L}},\tau_{\mathsf{U}}]$ that is bounded away from both 0 and 1 (i.e., $0<\tau_{\mathsf{L}} < \tau_{\mathsf{U}} < 1$). In contrast, our analysis critically requires careful treatment of extreme quantiles, which may explode as $\tau$ approaches 0 or 1 (Figure~\ref{invcdf}), to guarantee the convergence of the entire CDF of $\hat{Y}(\bx)$, including both its middle and tail parts. We develop a novel partitioning scheme that rigorously handles both middle and tail regions of the distribution, enabling a uniform convergence result.

Our third contribution concerns practical implementation. We provide guidance on choosing the number of quantile levels $m$ and address the quantile crossing issue. We recommend setting \( m = O(\sqrt{n}) \), balancing the convergence rate of \( \hat{Q}(\tau|\bx) \) in the main region for estimation error (governed by \( O_{\mathbb{P}}(1/\sqrt{n}) \) ) and for interpolation error (governed by \( O(1/m) \) ) with computational efficiency, as supported by our convergence analysis and numerical experiments. To handle quantile crossing, where estimated quantiles violate monotonicity, we leverage results from \cite{neocleous2008monotonicity}, showing that \( m = O(\sqrt{n}) \) ensures crossing frequency diminishes as \( n \) grows. For strict monotonicity, QRGMM can be paired with the rearrangement method of \citet{chernozhukov2010quantile} as a post-processing step with minimal overhead.

Finally, we extend QRGMM to improve its practical applicability. First, we show how to adapt it to handle \emph{multidimensional outputs} through sequential one-dimensional modeling. Second, we incorporate \emph{neural network-based quantile regression} to allow nonlinear quantile function approximation.

We note that quantile regression has been used in machine learning for building generative models—for example, in implicit quantile networks (IQN) for reinforcement learning \citep{dabney2018implicit} and PixelIQN for image generation \citep{ostrovski2018autoregressive}. However, these works primarily focus on empirical performance without theoretical convergence guarantees, which are essential for simulation-based decision-making. Our work bridges this gap by combining quantile function modeling with rigorous convergence analysis. %

The rest of the paper is organized as follows. Section~\ref{sec:formulation} formulates the generative metamodeling problem. Section~\ref{sec:QRGMM} introduces QRGMM. Section~\ref{sec:asy} establishes the convergence-in-distribution of QRGMM. Section~\ref{sec_choicem} discusses the choice of $m$. 
Section~\ref{sec:extension} provides extensions of QRGMM to permit multidimensional simulation outputs and the use of neural networks. Section~\ref{sec:num} presents comprehensive numerical experiments, with comparisons to widely used generative models. Section~\ref{sec:conclusions} draws conclusions, and the e-companion provides technical proofs and supplementary materials.

\section{Problem Formulation}\label{sec:formulation}

Consider a simulator designed to model a complex stochastic system for facilitating decision-making. The decision-maker provides input to this simulator in two parts: observable covariates $\bx$ and decision variables $\bd$. The stochastic system incorporates quantities beyond the full control of the decision-maker, represented by a random vector $\bxi$. The distribution of $\bxi$ may be affected by either $\bx$ or $\bd$. 
The simulator's output represents quantities indicating the performance of the stochastic system and is \emph{random} due to the presence of $\bxi$. Although lacking an analytical form, the simulator can be conceptualized as a \emph{deterministic} function $\Phi$ that maps the input and the random vector $\bxi$ to the output $\Phi((\bx,\bd), \bxi)$. For example, in a queueing system simulator, $\bx$ may include the arrival and service rates, $\bd$ may represent the number of servers, and $\bxi$ corresponds to the random variables involved in the queueing system, such as the inter-arrival times and service times. The simulator's output may be an overall cost accounting for both the staffing cost and the cost associated with service quality, such as average waiting time.

In this paper, we assume the decision variables $\bd$ are given and fixed, focusing on fast generation of approximate samples from the conditional distribution of the simulator's output given the covariates. To clarify this objective, we define the output as\[
Y(\bx) \coloneqq \Phi((\bx, \bd), \bxi). 
\]
Nevertheless, the developed methodology can be easily extended to learning the conditional distribution of $\Phi((\bx,\bd), \bxi)$ given both $\bx$ and $\bd$, and generating samples from this learned distribution.

In practice, simulators are often computationally intensive. Given the covariates $\bx$, generating sufficient samples of $Y(\bx)$ to achieve reliable statistical inference for informed decision-making can be time-consuming. To address this efficiency issue, metamodeling techniques are widely employed. 
Suppose the decision-maker is interested in the expected value of $Y(\bx)$. Given a simulation dataset $\{(\bx_i,y_i)\}_{i=1}^{n}$ with $y_i = Y(\bx_i)$, one can build a metamodel---such as linear regression, stochastic kriging, or neural networks---to capture the relationship between $\bx$ and $\mathbb{E}[Y(\bx)]$  \citep{barton2020tutorial}. This metamodel is \emph{predictive} in that it can directly predict the value of $\mathbb{E}[Y(\bx)]$ for any value of $\bx$ without running the simulator to collect samples of $Y(\bx)$.
In addition to the expectation, one can also build predictive metamodels for other summary statistics of $Y(\bx)$, such as tail probabilities \citep{jiang2020online} and quantiles \citep{chen2016efficient}.

A subtle yet significant limitation of these ``predictive'' metamodeling techniques is that the prediction objective---namely, the summary statistic of $Y(\bx)$---must be predetermined before constructing the metamodel. 
However, this requirement can be problematic for real-time decision-making. 
In dynamic environments, the relevant quantity for the decision-maker may fluctuate based on observed covariates or changing circumstances. 
Moreover, complex decisions often require balancing multiple objectives
whose relative importance may shift based on real-time information.

In this paper, we propose a new metamodeling framework to address the limitations of traditional predictive metamodels. We introduce \emph{generative metamodeling}, designed to meet the dynamic requirements of real-time decision-making. This approach accommodates scenarios where objectives may change instantaneously upon observing new covariates and where multiple objectives need to be addressed simultaneously. 

Specifically, we aim to construct a metamodel $\hat{Y}(\bx)$ that matches the distribution of the simulator's output $Y(\bx)$, rather than any specific summary statistic of $Y(\bx)$, denoted by $\mathcal{S}(Y(\bx))$. 
Hence, a generative metamodel is a \emph{random} function of the covariates, in contrast to a traditional (predictive) metamodel, which is a \emph{deterministic} function intended to approximate $\mathcal{S}(Y(\bx))$.

To be suitable for real-time decision-making, $\hat{Y}(\bx)$ should satisfy the following two criteria:
\begin{enumerate}[label=(\roman*)]
	\item Its distribution is approximately identical to that of $Y(\bx)$ for any $\bx$.
	\item Samples of $\hat{Y}(\bx)$ are significantly faster to generate than samples of $Y(\bx)$ for any $\bx$.
\end{enumerate}
Evidently, with a generative metamodel, one can approximately compute any summary statistic of $Y(\bx)$ in real time by quickly generating samples of $\hat{Y}(\bx)$.

\begin{remark}
Experimental design (i.e., how to design a simulation experiment to acquire the dataset) is crucial for developing high-quality metamodels \citep{barton2020tutorial}. 
In this paper, however, we proceed under the premise that a dataset $\{(\bx_i,y_i)\}_{i=1}^{n}$ already exists. This dataset may have been collected through a well-planned experimental design or could simply be a record of past experiments. We focus on how to build a generative metamodel given this dataset, leaving the study of experimental design for future research.

\end{remark}

\section{Quantile-Regression-Based Generative Metamodeling}\label{sec:QRGMM}

Given the variety of deep generative models for image generation \citep{Murphy23Probabilistic}, we emphasize that generative metamodeling has a distinct technical focus. 
We prioritize (i) distributional proximity between the generative metamodel and the simulator's output, and (ii) fast generation speed. 
In contrast, image generation models often aim to create visually appealing, diverse, and realistic-looking images for human perception. 

This paper focuses on continuous covariates $\bx$ and one- or low-dimensional output $Y(\bx)$, whereas image generation typically involves categorical covariates and high-dimensional output.
These cases are far from trivial. We seek theoretical characterization of the generative metamodel's distributional convergence, requiring careful analysis of tail behavior---an aspect uncommon in image generation.

\subsection{Framework}

Assume that $\bx \in \mathcal{X}\subset \mathbb{R}^p$ is a $p$-dimensional vector  and $Y(\bx) \in \mathbb{R}$ is a one-dimensional continuous random variable. (Section~\ref{sec:extension} discusses the case where $Y(\bx)$ is a random vector.)
Let
\[F_Y(y|\bx)\coloneqq \mathbb{P}(Y(\bx) \leq y|\bx)\quad\mbox{and}\quad 
F_Y^{-1}(\tau|\bx) \coloneqq \inf\{y: F_Y(y|\bx) \geq \tau \}
\]
denote the conditional CDF and conditional quantile function of $Y(\bx)$ given $\bx$, respectively.
It is well known that if $U \sim \mathsf{Unif}(0,1)$, then $F_Y^{-1}(U|\bx)$ has the same distribution as $Y(\bx)$, so we may use the former to generate samples of the latter.

However, $F_Y^{-1}(\tau|\bx)$ is generally unknown for simulators of complex systems.
In the following, we propose a framework (Algorithm~\ref{alg:gmm}) based on quantile regression to first learn this conditional quantile function using an offline simulation dataset, and then generate samples using the learned function to approximate the distribution of $Y(\bx)$.

In the offline stage, we specify a quantile regression model $Q(\tau|\bx)$. We focus on linear models for theoretical development, with neural network models discussed in Section~\ref{sec:extension}. For other options, see \cite{torossian2020review}.
Next, we discretize the interval $(0,1)$ into an equally-spaced grid $\{\tau_j \coloneqq j/m: j=1,\ldots,m-1\}$ for some integer $m$. Finally, we fit the model to the dataset $\{(\bx_i,y_i)\}_{i=1}^{n}$ for each $\tau_j$, yielding $\tilde{Q}(\tau_j|\bx)$ to predict $F^{-1}_Y(\tau_j|\bx)$ for any $\bx$.

In the online stage, after observing the covariates $\bx=\bx^*$,  we generate a sample $\{u_k: k=1,\ldots,K\}$ from $\mathsf{Unif}(0,1)$. For each $u_k$, we approximate $F^{-1}_Y(u_k|\bx^*)$ by $\hat{Q}(u_k|\bx^*)$, where $\hat{Q}(\tau|\bx)$ denotes a linear interpolator of the set of the predicted $\tau_j$-quantiles $\{(\tau_j, \tilde{Q}(\tau_j|\bx)):j=1,\ldots,m-1 \}$: 
\begin{equation} \label{eq:lin-interpo}
\hat{Q}(\tau|\bx) \coloneqq \left\{
\begin{array}{ll}
    \tilde{Q}(\tau_1|\bx), & \mbox{ if } \tau<\tau_{1}, \\
    \tilde{Q}(\tau_j|\bx)+m(\tau-\tau_j)\left[\tilde{Q}(\tau_{j+1}|\bx)-\tilde{Q}(\tau_j|\bx)\right], & \mbox{ if }
    \tau \in[\tau_j,\tau_{j+1}),\ j=1,\ldots,m-2, \\
    \tilde{Q}(\tau_{m-1}|\bx), & \mbox{ if } \tau\geq\tau_{m-1}. 
\end{array}\right.
\end{equation}
In particular, $\hat{Q}(\tau_j|\bx) = \tilde{Q}(\tau_j|\bx)$ for all $j=1,\ldots,m-1$. 
The generative metamodel then outputs $\{\hat{Y}_k(x^*)\coloneqq \hat{Q}(u_k|\bx^*):k=1,\ldots,K \}$ as an approximate sample of $Y(\bx)$.

\begin{algorithm}
	\caption{Quantile-Regression-Based Generative Metamodeling}\label{alg:gmm}
	\begin{algorithmic}
		\State \textbf{Offline Stage:}
		\begin{itemize}
		\item Collect a dataset $\{(\bx_i,y_i):i=1,\ldots,n\}$, and choose a quantile regression model $Q(\tau|\bx)$.
		\item Choose a positive integer $m$ and create the grid $\{\tau_j \coloneqq j/m: j=1,\ldots,m-1\}$.
		\item For each $j=1,\ldots,m-1$, fit the model with quantile level $\tau_j$ to the dataset, yielding $\tilde{Q}(\tau_j|\bx)$. 
		\end{itemize}			
		\State \textbf{Online Stage:} 
		\begin{itemize}
		\item Observe the covariates $\bx=\bx^*$.
		\item Generate $\{u_k:k=1,\ldots,K\}$ independently from $\mathsf{Unif}(0,1)$.
		\item Output: $\{\hat{Y}_k(x^*)\coloneqq \hat{Q}(u_k|\bx^*): k = 1,\ldots,K\}$, where $\hat{Q}(\tau|\bx)$ is the linear interpolator~\eqref{eq:lin-interpo}. 		
		\end{itemize}
	\end{algorithmic}
\end{algorithm}

\begin{remark}
The linear interpolation allows for fast approximation of $F^{-1}_Y(u_k|\bx^*)$ after observing $\bx^*$ in real time. Most quantile regression models require $\tau$ before fitting the model to data. Without interpolation, one would need to run quantile regression $K$ times in the online stage, once for each $u_k$. This process is computationally intensive and unsuitable for real-time decision-making.
\end{remark}

\begin{remark}
When generating realizations of $Y(\bx)$, capturing the tail behaviors of its conditional quantile function (as $\tau$ approaches 0 or 1) is generally challenging. In the QRGMM framework, we introduce the parameter $m$ to control learning fidelity and handle tails. As $m$ increases, the endpoints of the grid of quantile levels extend deeper into the tails (i.e., $\tau_1 \to 0$ and $\tau_{m-1} \to 1$). Understanding how $m$ should depend on the data size $n$ to ensure the distributional convergence of QRGMM-generated samples is critical. We address this important issue in Sections \ref{sec:asy}. 
\end{remark}

\subsection{Linear Quantile Regression Model}

The linear quantile regression model $Q(\tau|\bx) = \bbeta(\tau)^\intercal\bx$ is among the most widely used in practice. Here, $\bbeta(\tau)$ denotes the vector of linear coefficients that depend on $\tau$. 
We assume the conditional quantile function of $Y(\bx)$ is correctly specified by this model, and it is continuous: 
\begin{assumption}\label{ass:cons1} $F^{-1}_Y(\tau|\bx)=\bbeta(\tau)^\intercal\bx$	for all $\tau\in(0,1)$, and $\bbeta(\tau)\in\mathbb{R}^p$ is continuous in $\tau$.
\end{assumption}

An important special case of Assumption \ref{ass:cons1} is the \emph{location-scale shift} model:
\[
Y(\bx)=\balpha^\intercal\bx+(\bgamma^\intercal\bx) \varepsilon,
\]
where $\balpha$ and $\bgamma$ are constant vectors, and $\varepsilon$ is a one-dimensional random variable. Under this model, 
$
F^{-1}_Y(\tau|\bx)=\balpha^\intercal\bx+\bgamma^\intercal\bx F_{\varepsilon}^{-1}(\tau)
$
for all $\tau\in(0,1)$, where $F_{\varepsilon}^{-1}$ is the $\tau$-quantile of $\varepsilon$. Thus, the location-scale shift model satisfies Assumption~\ref{ass:cons1}. This holds regardless of the distribution of $\varepsilon$.

The linear quantile regression model can be generalized to a nonlinear model by incorporating basis functions: 
$Q(\tau|\bx) = \bbeta(\tau)^\intercal \bb(\bx)$, 
where $\bb(\bx)$ represents a vector of basis functions such as polynomial or radial basis functions. 
The theoretical results in this paper can be readily extended to this nonlinear case, and thus we focus on the linear model.

Under Assumption \ref{ass:cons1}, given a dataset $\{(\bx_i,y_i)\}_{i=1}^{n}$ and any quantile level $\tau\in(0,1)$, we can estimate $\bbeta(\tau)$ by solving the minimization problem
\begin{equation}\label{eq:optqr}
\min_{\bbeta\in{\mathbb R}^p} \sum_{i=1}^n\rho_\tau (y_i-\bbeta^\intercal\bx_i),
\end{equation}
where 
$\rho_\tau(u)\coloneqq (\tau-\mathbb{I}\{u\le 0\})u$ is the \emph{pinball loss function} and $\mathbb{I}\{\cdot\}$ is the indicator function. 
This loss function generalizes the absolute value loss function used for median regression.
Note that $\rho_\tau(u) = \tau u$ if $u\geq 0$ and $\rho_\tau(u) = (1-\tau) |u|$ if $u < 0$.
Thus, this loss asymmetrically penalizes under- and overestimation errors when estimating $y_i$ with $\bbeta^\intercal \bx_i$.
It assigns a weight of $\tau$ to underestimation errors $(y_i - \bbeta^\intercal \bx_i)$
and $(1-\tau)$ to overestimation errors $(\bbeta^\intercal \bx_i - y_i)$.
This asymmetry allows the quantile regression to target specific quantiles of the conditional distribution.
For instance, if underestimation costs three times more than overestimation (i.e., $\tau/(1-\tau) = 3$),
then the solution to problem~\eqref{eq:optqr}, which balances the two types of estimation costs, ensures the probability of underestimation is three times less than the probability of overestimation, thus leading to a $75\%$-quantile.

To adopt the linear quantile regression model in the QRGMM framework,
in the offline stage,
we solve problem~\eqref{eq:optqr} with $\tau = \tau_j$ for each $j=1,\ldots,m-1$ to obtain the solution
\begin{equation}\label{eq:tilde-beta}
\tilde{\bbeta}({\tau_j}) \coloneqq \argmin_{\bbeta\in\mathbb{R}^p} \sum_{i=1}^n\rho_{\tau_j}(y_i-\bbeta^\intercal\bx_i).
\end{equation}
This can be done via reformulating the minimization problem as a linear program, which can be solved efficiently \citep{koenker2017handbook}.
Then, we define the interpolated coefficient vector $\hat{\bbeta}(\tau)$ as
\begin{equation} \label{eq:lin-interpo-beta}
\hat{\bbeta}(\tau) \coloneqq \left\{
\begin{array}{ll}
    \tilde{\bbeta}(\tau_1), & \mbox{ if } \tau<\tau_{1}, \\
    \tilde{\bbeta}(\tau_j) + m(\tau-\tau_j)\left[\tilde{\bbeta}(\tau_{j+1}) -\tilde{\bbeta}(\tau_j) \right], & \mbox{ if }
    \tau \in[\tau_j,\tau_{j+1}),\ j=1,\ldots,m-2, \\
    \tilde{\bbeta}(\tau_{m-1}), & \mbox{ if } \tau\geq\tau_{m-1}. 
\end{array}\right.
\end{equation}
The linear interpolator~\eqref{eq:lin-interpo} becomes $\hat{Q}(\tau|\bx) = \hat{\bbeta}(\tau)^\intercal \bx$.

In the online stage, for any given covariates $\bx = \bx^*$,
we generate $\{\hat{Y}_k(\bx^*):k=1,\ldots,K\}$ to approximate the distribution of $Y(\bx^*)$, where $\hat{Y}_k(\bx^*) = \hat{\bbeta}(u_k)^\intercal \bx^*$.

\section{Convergence in Distribution}\label{sec:asy}

In this section, we prove that for any fixed $\bx^*\in \mathcal{X}$, the generative metamodel $\hat{Y}(\bx^*)\coloneqq \hat{Q}(U|\bx^*)$ with the linear quantile regression model converges to the simulator's output $Y(\bx^*)$ in distribution as $n$, the size of the offline simulation dataset, grows.
As the goal of generative metamodeling is fast generation of samples that can be used to estimate any summary statistic of $Y(\bx^*)$,
this convergence in distribution provides the validity of QRGMM. Due to space limitations, all the proofs in this section are placed in the e-companion.

A technical challenge in proving $\hat{Y}(\bx^*) \Rightarrow Y(\bx^*)$, where ``$\Rightarrow$'' denotes convergence in distribution, lies in careful analysis of the tail behaviors of the distributions of $\hat{Y}(\bx^*)$ and $Y(\bx^*)$. 
To appreciate this challenge, 
note that the distribution of $Y(\bx^*)$ is fully characterized by its conditional quantile function $F^{-1}_Y(\tau|\bx^*)$. 
As shown in Figure~\ref{invcdf}, $F^{-1}_Y(\tau|\bx^*)$ may go to infinity  very quickly as $\tau$ approaches 0 or 1 if $Y$ has an infinity support, 
causing difficulties in bounding the approximation errors of $\hat{Q}(\tau|\bx^*)$ relative to $F^{-1}_Y(\tau|\bx^*)$.

\begin{figure}[t]	\FIGURE{\includegraphics[width=0.6\textwidth]{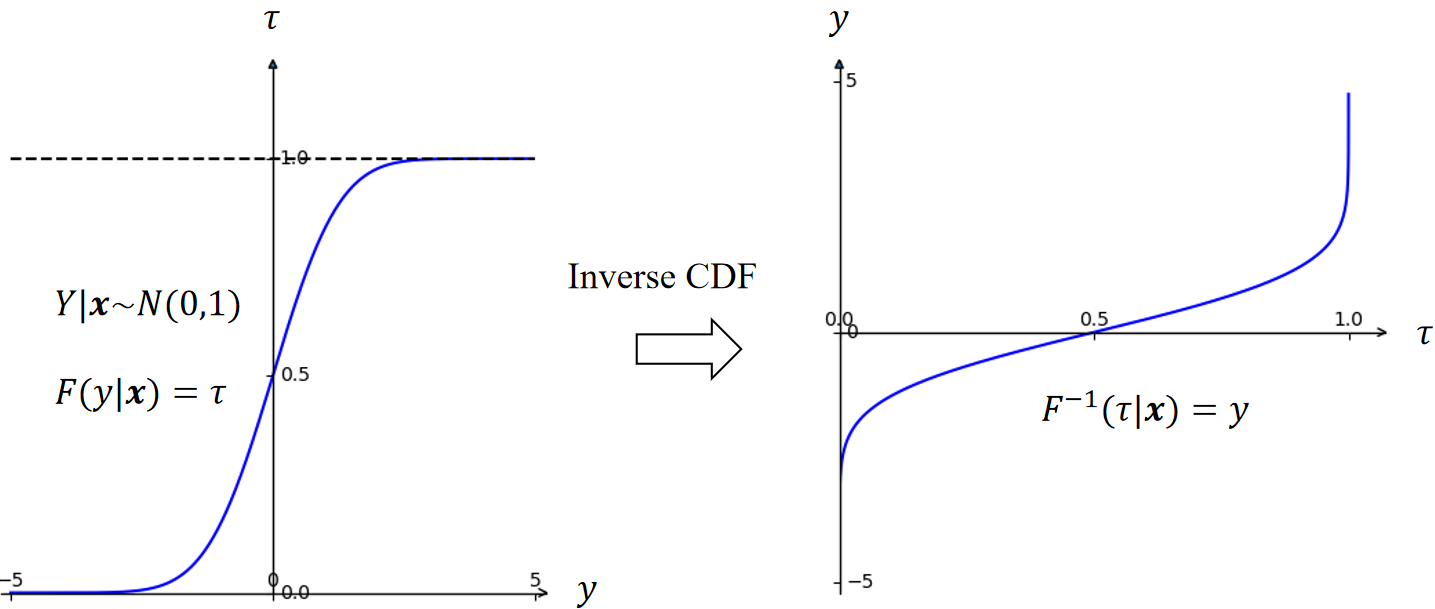}}
{Cumulative Distribution Function and Quantile Function of $Y(\bx)$. 	\label{invcdf}}
{}
\end{figure}

This difficulty also appears in quantile regression literature. 
Typically, $\tau$ is assumed fixed or restricted to $[\tau_{\mathsf{L}},\tau_{\mathsf{U}}]$, where $0<\tau_{\mathsf{L}}<\tau_{\mathsf{U}}<1$ \citep{koenker2002inference}. 
This approach effectively excludes the tails from asymptotic analysis. 
To overcome this limitation, we develop a partition scheme. 
This scheme extends prior quantile regression results from $[\tau_{\mathsf{L}},\tau_{\mathsf{U}}]$ to the entire interval $(0,1)$, allowing us to prove QRGMM's convergence in distribution.

\subsection{Preliminaries}

\begin{assumption}\label{ass:cons2}	
	\begin{enumerate}[label=(2.\alph*), left=0pt]
		\item \label{ass:cons2a}  $F_Y(\cdot|\bx)$ is absolutely continuous with  a density function $f_Y(\cdot|\bx)$ on a support that is common for all $\bx\in\mathcal{X}$. 
		\item \label{ass:cons2b} For any $0<\tau_{\mathsf{L}}<\tau_{\mathsf{U}}<1$, 
 $f_Y(\cdot|\bx)$ is uniformly bounded above and bounded below away from $0$,  for all $\bx\in\mathcal{X}$ and all $\tau\in [\tau_{\mathsf{L}}, \tau_{\mathsf{U}}]$  in the following sense:
		\[
\inf_{\tau\in[\tau_{\mathsf{L}},\tau_{\mathsf{U}}]}\inf_{\bx\in\mathcal{X}} f_Y(F_Y^{-1}(\tau|\bx)|\bx) > 0 \quad\mbox{and}\quad \sup_{\tau\in[\tau_{\mathsf{L}},\tau_{\mathsf{U}}]}\sup_{\bx\in\mathcal{X}} f_Y(F_Y^{-1}(\tau|\bx)|\bx)<\infty.
		\]
	\end{enumerate} 
\end{assumption}

\begin{assumption}\label{ass:cons3}
	\begin{enumerate}[label=(3.\alph*), left=0pt]
		\item There exist a positive definite matrix $D_0$ and a family of positive definite matrices
		$\{D_1(\tau):\tau\in(0,1)\}$ such that
		\[
		\lim_{n\to\infty}\frac{1}{n}\sum_{i=1}^n\bx_i\bx_i^\intercal=D_0 \quad\text{and}\quad\lim_{n\to\infty}\frac{1}{n}\sum_{i=1}^nf_Y(F^{-1}_Y(\tau|\bx_i)|\bx_i)\bx_i\bx_i^\intercal=D_1(\tau)
		\] 
  almost surely. Moreover, for any $0<\tau_L<\tau_U<1$, the second convergence holds uniformly in $\tau$ over $\tau\in[\tau_L,\tau_U]$. \label{ass:cons3a}	
		\item $\max_{1\le i\le n}\|\bx_i\|=o(\sqrt{n})$. \label{ass:cons3b}	
		
	\end{enumerate} 
\end{assumption}

These assumptions are standard for asymptotic analysis of quantile regression \citep{koenker2005}.  
Assumption~\ref{ass:cons2} pertains to the conditional distribution of $Y(\bx)$ given $\bx$. 
Condition~\ref{ass:cons2a} imposes a smoothness requirement on the conditional CDF of $Y(\bx)$. 
Condition~\ref{ass:cons2b} stipulates that the density function $f_Y(y|\bx)$ is uniformly bounded both above and below for all $\bx\in\mathcal{X}$ and for all $y$ between the $\tau_{\mathsf{L}}$- and $\tau_{\mathsf{U}}$-quantiles of $Y(\bx)$. These conditions are typically satisfied, particularly when $\mathcal{X}$ is bounded.
Assumption~\ref{ass:cons3} concerns the design points $\{\bx_1,\ldots,\bx_n\}$, which represent the observed covariates in the offline simulation dataset.
When $\mathcal{X}$ is bounded, conditions \ref{ass:cons3a} and \ref{ass:cons3b} are satisfied.

In addition to the standard $o(\cdot)$ and $O(\cdot)$ notation, we use their stochastic counterparts $o_{\mathbb P}(\cdot)$ and $O_{\mathbb P}(\cdot)$, which are common in statistics literature. 
Here, $o_{\mathbb P}(1)$ denotes a sequence of random vectors converging to zero in probability, while $O_{\mathbb P}(1)$ represents a sequence of random variables bounded in probability. 
Analogous to $o(r_n)$ and $O(r_n)$, the expressions $o_{\mathbb P}(r_n)$ and $O_{\mathbb P}(r_n)$ can be interpreted as sequences divided by $r_n$, following to the definitions of $o_{\mathbb P}(1)$ and $O_{\mathbb P}(1)$ respectively.

\begin{lemma}[Bahadur Representation]\label{lemma:brown} 
For any $\tau$, 
let $\tilde{\bbeta}(\tau)$ denote the linear quantile regression estimator of $\bbeta(\tau)$ by solving problem~\eqref{eq:optqr}. 
Fix any interval $[\tau_{\mathsf{L}},\tau_{\mathsf{U}}]\subset(0,1)$ with $0<\tau_{\mathsf{L}}<\tau_{\mathsf{U}}<1$. 
Under Assumptions~\ref{ass:cons1}--\ref{ass:cons3}, 
\begin{equation}\label{eqn:br1}
\sqrt{n}\left[\tilde{\bbeta}(\tau)-\bbeta(\tau)\right]=D^{-1}_1(\tau)W_n(\tau)+o_{\mathbb P}(1),
\end{equation}
as $n\to\infty$, uniformly for $\tau\in[\tau_{\mathsf{L}},\tau_{\mathsf{U}}]$. 
Here, \begin{equation*}%
		W_n(\tau) \coloneqq \frac{1}{\sqrt{n}}\sum_{i=1}^n\bx_i\psi_\tau\left(y_i-\bbeta(\tau)^\intercal\bx_i\right) \Rightarrow \mathcal{N}(0,\tau(1-\tau)D_0), 
	\end{equation*} 
 as $n\to\infty$,  where 
 $\psi_\tau(u)= \tau-\mathbb{I}\{u\le 0\}$ and $\mathcal{N}(\mu, \sigma^2)$ denotes a normal distribution with mean $\mu$ and variance $\sigma^2$. 
 Moreover, 
 \begin{equation}\label{eqn:br2}	\sqrt{n}D_0^{-1/2}D_1(\tau)\left[\tilde{\bbeta}(\tau)-\bbeta(\tau)\right]\Rightarrow\bB(\tau),
	\end{equation}
 as $n\to\infty$, 
 uniformly for $\tau\in[\tau_{\mathsf{L}},\tau_{\mathsf{U}}]$, where $\mathbf{B}$ is a $p$-variate process of independent Brownian bridges. 
\end{lemma}

The Bahadur representation, originally developed for quantile estimation \citep{bahadur1966note}, is crucial in understanding the asymptotic behaviors of various quantile estimators \citep{sun2010asymptotic,chu2012confidence}. Lemma~\ref{lemma:brown}, which can be found in \citet[Chapter~4.3]{koenker2005}, extends this representation to quantile regression and applies it to a range of $\tau$ values rather than a single fixed value. The convergence result \eqref{eqn:br2} is particularly critical to our analysis, as we focus on the convergence of $\hat{Q}(\tau|\bx)$ to $F^{-1}_Y(\tau|\bx)$ for all $\tau\in(0,1)$, not just a fixed $\tau$.

Assumptions~\ref{ass:cons2b} and \ref{ass:cons3a} imply $\sup_{\tau\in[\tau_{\mathsf{L}},\tau_{\mathsf{U}}]}|D^{-1}_1(\tau)|<\infty$. 
Hence, the convergence \eqref{eqn:br2} readily yields 
the uniform convergence of $\tilde{\bbeta} (\tau_j)$ for all $\tau_j\in[\tau_{\mathsf{L}},\tau_{\mathsf{U}}]$, 
as demonstrated in the Proposition \ref{pro:asym1}.

\begin{proposition}\label{pro:asym1}
	Fix any interval $[\tau_{\mathsf{L}},\tau_{\mathsf{U}}]\subset(0,1)$ with $0<\tau_{\mathsf{L}}<\tau_{\mathsf{U}}<1$. 
Under Assumptions~\ref{ass:cons1}--\ref{ass:cons3}, 
\[
\max_{\substack{\tau_j \in\left[\tau_{\mathsf{L}}, \tau_{\mathsf{U}}\right] \\ 1 \leq j \leq m-1}}\left\|\tilde{\bbeta}(\tau_j)-\bbeta\left(\tau_j\right)\right\| \stackrel{\mathbb{P}}{\rightarrow} 0,
\]
 as $n \rightarrow \infty$, where $\tau_j=j/m$ and ``$\stackrel{\mathbb{P}}{\rightarrow}$'' denotes convergence in probability.  
\end{proposition}

The Bahadur representation in Lemma~\ref{lemma:brown} and Proposition~\ref{pro:asym1} provide convergence results for quantile regression coefficients only on a given interval $[\tau_{\mathsf{L}},\tau_{\mathsf{U}}]$. This is insufficient for proving the distributional convergence of $\hat Y(\bx^*)$, as many $\tau_j$'s fall outside $[\tau_{\mathsf{L}},\tau_{\mathsf{U}}]$ when $m\to\infty$. New techniques are required. Before presenting them in Section~\ref{subsec:truncation}, we introduce an additional tool for addressing the distributional convergence of $\hat Y(\bx^*)$: the Portmanteau theorem, a well-known result for proving convergence in distribution; see, e.g., \citet[page~6]{van2000}.

\begin{lemma}[Portmanteau Theorem]\label{lemma:van} For any random vectors $X_{n}$ and $X$, the following statements are equivalent:
\begin{enumerate}[label=(\roman*)]
    \item $X_n\Rightarrow X$ as $n\to\infty$;
    \item $\mathbb{E} \left[h\left(X_{n}\right)\right] \rightarrow \mathbb{E} [h(X)]$ as $n\to\infty$ for all bounded Lipschitz continuous functions $h$.
	\end{enumerate}
\end{lemma}

The Portmanteau theorem offers a convenient approach to handling the tails of the distribution of $\hat{Y}(\bx^*) = \hat{Q}(U|\bx^*)$. While $\hat{Q}(\tau|\bx^*)$ may be unbounded and grow rapidly to infinity as $\tau$ approaches 0 or 1, this is no longer the case after applying the function $h$. The conditional quantile function of $h(\hat{Q}(\tau|\bx^*))$ is bounded, with growth rate limited by the Lipschitz constant. This property is crucial in proving the distributional convergence of QRGMM.

\subsection{Partition Scheme}\label{subsec:truncation}

To show $\hat{Y}(\bx^*) \Rightarrow Y(\bx^*)$, it is equivalent to proving 
$\hat{Q}(U|\bx^*) \Rightarrow F_Y^{-1}(U|\bx^*)$. 
According to the Portmanteau theorem, this 
is equivalent to showing 
\begin{equation}\label{eqn:port1}
\mathbb{E} \left[h\left(\hat{Q}(U|\bx^*)\right)\right]\to \mathbb{E} \left[h\left(F^{-1}_Y(U|\bx^*)\right)\right],
\end{equation} 
for all bounded Lipschitz continuous functions $h$.

We now introduce a partition scheme to facilitate the asymptotic analysis of $\hat{Q}(\tau|\bx^*)$ when $\tau$ approaches the tails of the distribution. 
Fix an arbitrary bounded Lipschitz continuous function $h$.
Let $M_h$ and $L_h$ denote the bound and the Lipschitz constant of this function, respectively. 
That is, 
$|h(y)|\le M_h$ and $|h(y)-h(y')|\le L_h|y-y'|$ for all $y$ and $y'$. 
Fix an arbitrary $\epsilon>0$ and define 
\begin{equation}\label{eqn:tau}
\tau_{\mathsf{L}}=\frac{\epsilon}{10\max(M_h,1)}\quad \mbox{and} \quad \tau_{\mathsf{U}} =1-\frac{\epsilon}{10\max(M_h,1)}.
\end{equation}

Our partition scheme divides the interval $(0,1)$ into five parts:
\begin{equation}\label{eqn:int}
I_1=(0,\tau_{\mathsf{L}}),\ I_2=\left[\tau_{\mathsf{L}},\tau_{\mathsf{L}}+\frac{1}{m}\right),\ I_3=\left[\tau_{\mathsf{L}}+\frac{1}{m},\tau_{\mathsf{U}}-\frac{1}{m}\right],\ I_4=\left(\tau_{\mathsf{U}}-\frac{1}{m},\tau_{\mathsf{U}}\right],\ I_5=(\tau_{\mathsf{U}},1).
\end{equation}
Among these five sub-intervals, $I_1$ and $I_5$ represent the tails, while $I_3$ is the middle interval.
Note that $I_3$ is a subset of $[\tau_{\mathsf{L}},\tau_{\mathsf{U}}]\cap \left[\frac{1}{m},1-\frac{1}{m}\right]$.
We define it this way to ensure that $I_3$ lies within $[\tau_1,\tau_{m-1}]$, the range of quantile levels used in the offline stage of the QRGMM framework.
The sub-intervals $I_2$ and $I_4$ are included for technical completeness and may be viewed as parts of the tails.

Clearly, to prove \eqref{eqn:port1}, it suffices to show that 
\begin{equation}\label{eqn:port2}
\mathbb{E}\left[ \left|h\left( \hat{Q}(U|\bx^*) \right) - h\left( F^{-1}_Y(U|\bx^*) \right)\right|\cdot \mathbb{I}\{U\in I_i\} \right]\le\frac{\epsilon}{5},
\end{equation}
for all $i=1,\ldots,5$. 
It is easy to show this inequality holds for $I_1$, $I_2$, $I_4$, and $I_5$.

\begin{lemma} \label{lemma:I1245} 
Fix $\bx^*\in\mathcal{X}$.
For any bounded Lipschitz continuous function $h$ and $\epsilon>0$, 
\[
\mathbb{E}\left[ \left|h\left( \hat{Q}(U|\bx^*) \right) - h\left( F^{-1}_Y(U|\bx^*) \right)\right|\cdot \mathbb{I}\{U\in I_i\} \right]\le{\frac{\epsilon}{5}},
\]
for all $i=1,2,4,5$ and $m\ge {10\max(M_h,1)/\epsilon}$.
\end{lemma}

For the middle interval $I_3$, we establish a stronger convergence result.

\begin{proposition}\label{pro:asym2}
    Suppose Assumptions~\ref{ass:cons1}--\ref{ass:cons3} hold. Then, for any $\bx^*\in\mathcal{X}$, 
    \begin{eqnarray}
        \lefteqn{ \sup_{\tau\in I_3}\left|\hat{Q}(\tau|\bx^*)-F^{-1}_Y(\tau|\bx^*)\right|  } \nonumber\\
        & \le & \max_{\substack{\tau_{j}\in[\tau_{\mathsf{L}},\tau_{\mathsf{U}}] \\ 1\le j\le {m-1}}}\|\tilde{\bbeta}(\tau_j)-\bbeta(\tau_j)\|\cdot\|\bx^*\| + \frac{1}{m}\left(\inf_{\tau\in[\tau_{\mathsf{L}},\tau_{\mathsf{U}}]}f_Y\left(F_Y^{-1}(\tau|\bx^*)|\bx^*\right)\right)^{-1}.  \label{pf:asym2}
    \end{eqnarray}
    Furthermore, as $n, m \rightarrow \infty$,
    \begin{equation}\label{eq:uniform-converg-I3}
        \sup_{\tau\in I_3}\left|\hat{Q}(\tau|\bx^*)-F^{-1}_Y(\tau|\bx^*)\right| \stackrel{\mathbb{P}}{\rightarrow} 0.
    \end{equation}
\end{proposition}

Proposition~\ref{pro:asym2}, particularly the inequality~\eqref{pf:asym2}, implies that
the approximation error of $\hat{Q}(\tau|\bx^*)$ relative to $F_Y^{-1}(\tau|\bx^*)$ is bounded, uniformly for $\tau\in I_3$, by the sum of two components.
The first component pertains to the \emph{estimation error} of $\tilde{\bbeta}(\tau_j)$ relative to $\bbeta(\tau_j)$ in the linear quantile regression model \eqref{eq:tilde-beta}.
The second component is the \emph{interpolation error} arising from the linear interpolation scheme \eqref{eq:lin-interpo-beta}, which constructs $\hat{\bbeta}(\tau)$ based on $\tilde{\bbeta}(\tau_j)$.
The estimation error diminishes as $n\to\infty$, while the interpolation error does likewise as $m\to\infty$, as ensured by Assumption \ref{ass:cons2b}.
This leads to the convergence of $\hat{Q}(\tau|\bx^*)$ to $F_Y^{-1}(\tau|\bx^*)$ uniformly for $\tau\in I_3$. 
With Proposition~\ref{pro:asym2}, we can establish that the inequality~\eqref{eqn:port2} holds for $I_3$. 

\begin{lemma}\label{lemma:I3} 
Fix $\bx^*\in\mathcal{X}$.
Suppose Assumptions~\ref{ass:cons1}--\ref{ass:cons3} hold. Then, 
for any bounded Lipschitz continuous function $h$ and $\epsilon>0$, there exist $M,N>0$ such that
    \[
        \mathbb{E}\left[ \left|h\left( \hat{Q}(U|\bx^*) \right) - h\left( F^{-1}_Y(U|\bx^*) \right)\right|\cdot \mathbb{I}\{U\in I_3\} \right]\le{\frac{\epsilon}{5}},
    \]
for all $m \geq M$ and $n\geq N$.
\end{lemma}

\subsection{Establishing the Convergence in Distribution}

Combining Lemmas~\ref{lemma:I1245} and \ref{lemma:I3} yields the convergence \eqref{eqn:port1},
which, by the Portmanteau theorem, establishes $\hat{Y}(\bx^*)\Rightarrow Y(\bx^*)$.
Additionally, we can establish a uniform convergence result for the CDF of $\hat{Y}(\bx^*)$.
The proof of this result is included in the e-companion.

\begin{theorem}\label{thm:weakconvergence}
	Under Assumptions~\ref{ass:cons1}--\ref{ass:cons3}, the following results hold:
	\begin{enumerate}[label=(1.\alph*), left=0pt]
		\item For any $\bx^*\in \mathcal{X}$, $\hat{Y}(\bx^*) \Rightarrow Y(\bx^*)$ as $n,m\to\infty$ . \label{thm:weakconvergence_a}	
		\item For any $\bx^*\in \mathcal{X}$,
		\[
		\sup_{y\in\left(-\infty,\infty\right)}\left|F_{\hat{Y}}(y|\bx^*)-F_Y(y|\bx^*)\right| \to 0,
            \]
		as $n,m\to\infty$, 
		where $F_{\hat{Y}}(y|\bx^*)$ is the conditional CDF of $\hat{Y}(\bx^*)$ given $\bx^*$. \label{thm:weakconvergence_b}	
	\end{enumerate} 

\end{theorem}

\section{Discussion on the Choice of $m$}\label{sec_choicem}

The uniform convergence of the distribution function of $\hat{Y}(\bx^*)$, as established in Theorem \ref{thm:weakconvergence}, validates the asymptotic correctness of the QRGMM algorithm for learning conditional distributions and generating conditional random variables. However, practical implementation requires guidance on selecting the number of quantile levels, $m$, relative to the sample size, $n$. This section provides some analysis for addressing this issue. Specifically, we analyze the rate of convergence in the middle interval and the quantile crossing phenomenon to justify the recommended choice of $m = O(\sqrt{n})$.

\subsection{Rate of Convergence in the Middle Interval}\label{m_choice_rate}

As convergence in distribution does not necessarily imply that the random variables ${\hat{Y}}(\bx^*)$ and $Y(\bx^*)$ are approaching each other as $n\to\infty$.
To understand this issue, we note that ${\hat{Y}}(\bx^*)=\hat{Q}(U|\bx^*)$ and $Y(\bx^*)=F^{-1}_Y(U|\bx^*)$ where $U\sim{\rm Unif}(0,1)$. Then, it is natural to examine the convergence behavior of $\hat{Q}(\tau|\bx^*)$ to $F^{-1}_Y(\tau|\bx^*)$ for any $\tau\in(0,1)$, effectively adopting a sample-path viewpoint by considering a fixed realization of $U$. This approach provides a more granular understanding of how ${\hat{Y}}(\bx^*)$ approximates $Y(\bx^*)$ as $n\to\infty$, and provide guidance for the choice of $m$.

In this context, we focus on the middle interval, i.e., $I_3$ in Equation (\ref{eqn:int}). This choice is motivated by both theoretical and practical considerations. From a theoretical standpoint, extreme quantile regression—i.e., when $\tau \to 0$ or $\tau \to 1$—is a highly complex issue and exhibits distinct asymptotic behavior due to the potential vanishing of the conditional density \citep{chernozhukov2005extremal}, we instead concentrate on the middle interval, where the conditional density is assumed to be bounded and well-behaved. From a practical data generation perspective, the middle interval is of primary importance because it typically captures the central behavior of the distribution, where most data points lie and where estimation is more robust due to sufficient data availability. In addition, reliable estimation of extreme quantiles typically requires far more data, which is often unavailable in practice. 

We now present our main results on the convergence rate of $\hat{Q}(\tau|\bx^*)$ in this region. We begin by explicitly deriving the rate of convergence of $\hat\bbeta_j$ within $[\tau_\ell,\tau_u]$, with proof in the e-companion.

\begin{proposition}\label{pro:asym3}
	Under Assumptions~\ref{ass:cons1}--\ref{ass:cons3}, 
	we have that as $n\to\infty$,  
	\[
	\max_{\substack{\tau_{j}\in[\tau_\ell,\tau_u] \\ 1\le j\le m-1}}\left\|\hat\bbeta_j-\bbeta(\tau_j)\right\|=O_{\mathbb P}\left(\frac{1}{\sqrt{n}}\right).
	\]
	
\end{proposition}

Then, combined with Proposition \ref{pro:asym2}, the rate of convergence of $\hat{Q}(\tau|\bx^*)$ for the middle interval can be easily obtained. We summarize the result in the following proposition.
\begin{proposition}\label{pro:asym4}
	Under Assumptions~\ref{ass:cons1}--\ref{ass:cons3}, for any given $\bx^*$, as $n, m\to\infty$, 
	\[
	\sup_{\tau\in[\tau_\ell+\frac{1}{m},\tau_u-\frac{1}{m}]}\left|\hat{Q}(\tau|\bx^*)-F^{-1}_Y(\tau|\bx^*)\right|=O_{\mathbb P}\left(\frac{1}{\sqrt{n}}\right)+O\left(\frac{1}{m}\right).
	\]
\end{proposition}

Proposition \ref{pro:asym4} reveals that the convergence rate of $\hat{Q}(\tau|\bx^*)$ in the middle interval is governed by the slower of the two terms: $O_{\mathbb{P}}(1/\sqrt{n})$ (estimation error) and $O(1/m)$ (interpolation error). To ensure that the interpolation error does not dominate, $m$ should grow at least as fast as $\sqrt{n}$. Thus, a practical choice is $m = O(\sqrt{n})$, such as $m = \sqrt{n}$, which balances the convergence rate of $O_{\mathbb{P}}(1/\sqrt{n})$ with the computational cost of fitting $m$ quantile regression models.

\subsection{Quantile Crossing}

Another critical consideration for selecting $m$ is the quantile crossing problem, where estimated conditional quantiles violate the monotonicity property ($\hat{Q}(u_1|\bx^*) \geq \hat{Q}(u_2|\bx^*)$ for $u_1 < u_2$). This issue often arises when running quantile regression for multiple quantile levels.

\subsubsection{Frequency of Quantile Crossing.}\label{subsec:crossing-freq}

\citet{neocleous2008monotonicity} proved that when fitting linear quantile regression with $m = O(\sqrt{n}/\log(n))$ quantile levels and using linear interpolation, conditional quantile estimates become strictly monotone with probability approaching one as $n\to\infty$, under mild conditions (see Lemma~\ref{lemma:neo} in the e-companion). This implies quantile crossing becomes increasingly rare with larger sample sizes. Our recommended choice of $m = O(\sqrt{n})$ in Section \ref{m_choice_rate} differs only by a $\log(n)$ factor. 
To verify that this choice maintains diminishing quantile crossing frequency as $n\to\infty$, we conduct numerical experiments using Test Problem 1 from Section \ref{sec:synthetic}.

For a given $\bx^*$, $n$, and $m$, we measure quantile crossing by comparing the original estimated quantiles $\{\hat{Q}(\tau_j|\bx^*):j=1,\ldots,m-1\}$ with their sorted sequence $\{q_j:j=1,\ldots,m-1\}$. When no crossing occurs, $\hat{Q}(\tau_j|\bx^*) = q_j$ for all $j$. We define the crossing frequency as $(m-1)^{-1}\sum_{j=1}^{m-1} \mathbb{I}\{\hat{Q}(\tau_j|\bx^*)\neq q_j\}$. 
Using $m = c\sqrt{n}$ for $c = 1, 5, 10$ and varying $n$ from $10^2$ to $10^4$, we compute crossing frequencies across ten replications, with results shown in Figure~\ref{fig:quantilecrossingrate_ngrow}. The results confirm that quantile crossing becomes negligible when $n$ exceeds $10^3$ for all the three values of $c$. Even in the most demanding case of $c=10$, which yields $m=1000$ quantile levels when $n=10^4$, the crossing frequency is virtually zero.

\begin{figure}[ht]
\FIGURE{\includegraphics[width=0.6\textwidth]
	{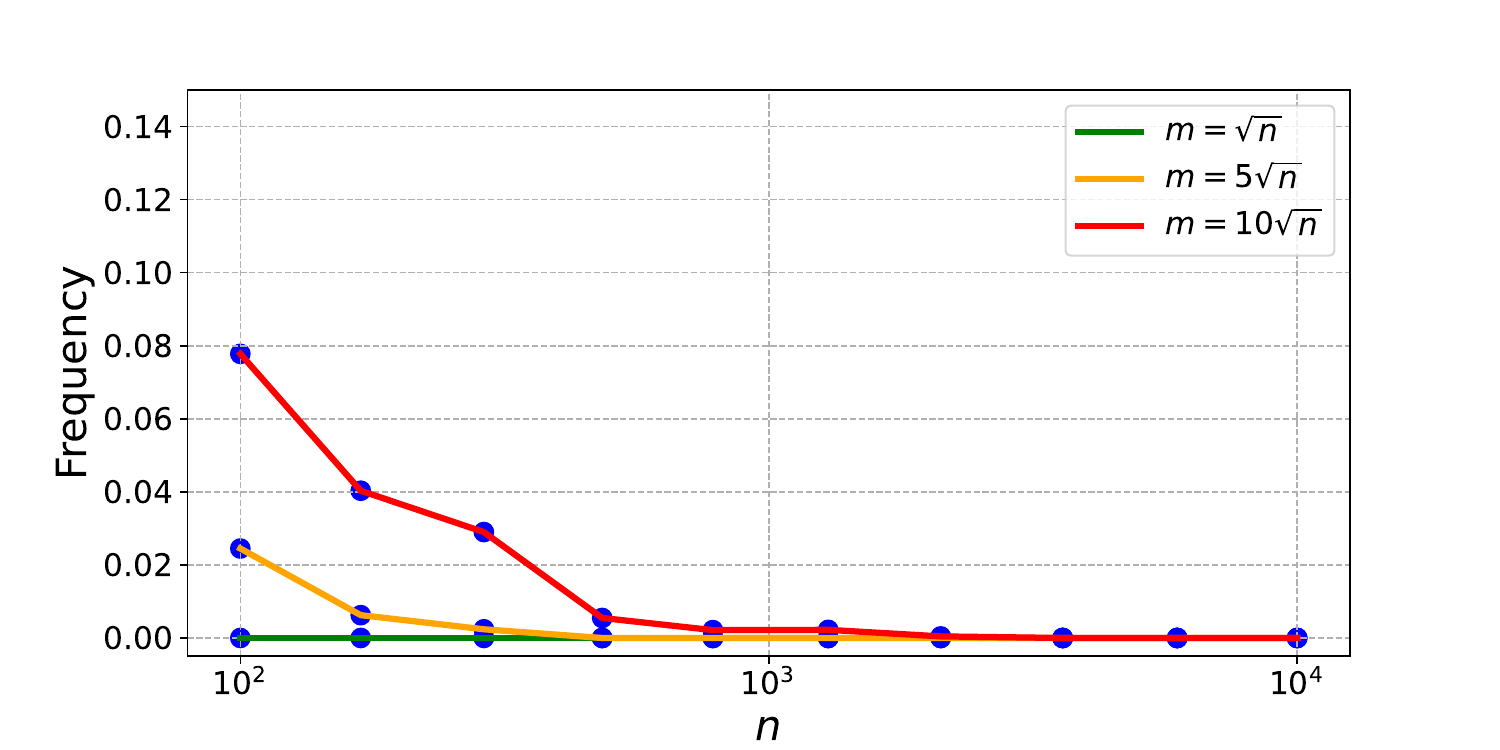}}%
	{Quantile Crossing Frequency with $m=O(\sqrt{n})$. \label{fig:quantilecrossingrate_ngrow}}
	{The results are calculated for Test Problem 1 with $\bx^*=(1,6,1,2)$ based on 10 replications.}
\end{figure}

 \subsubsection{Post-processing with Rearrangement.}

 When quantile crossing occurs, we can apply \cite{chernozhukov2010quantile}'s rearrangement method to restore monotonicity. 
 This method is straightforward to implement, and works as a model-independent post-processing step. This independence makes it particularly suitable for the QRGMM framework, which we design to accommodate various quantile regression specifications, unlike alternative methods that enforce monotonicity via model-specific constraints or penalties during training \citep{bondell2010noncrossing,dai2022non}.

 Our generative metamodel is $\hat{Y}(\bx)=\hat{Q}(U | \bx)$, where $U \sim \mathsf{Unif}(0,1)$. 
 Let $\hat{Q}^{\mathsf{R}}(\tau|\bx)$ denote the conditional quantile function of $\hat{Y}(\bx)$ given $\bx$, i.e., $\hat{Q}^{\mathsf{R}}(\tau|\bx)\coloneqq  \inf \{y: F_{\hat{Y}}(y | \bx) \geq \tau\}$, where $F_{\hat{Y}}(y|\bx)$ is the conditional CDF of $\hat{Y}(\bx)$ given $\bx$. 
 The rearrangement method transforms $\hat{Q}(\tau|\bx)$, which is potentially non-monotone in $\tau$, into $\hat{Q}^{\mathsf{R}}(\tau|\bx)$, which is guaranteed to be monotonically increasing in $\tau\in(0,1)$ and coincides with $\hat{Q}(\tau|\bx)$ when the latter is already monotone.
 
 The rearrangement method is incorporated into the QRGMM framework as follows. After obtaining the estimates $\{\tilde{Q}(\tau_1|\bx), \ldots, \tilde{Q}(\tau_{m-1}|\bx)\}$ from the quantile regression model, we set $\tilde{Q}^{\mathsf{R}}(\tau_j|\bx)$ as the $j$-th order statistic of these estimates, effectively sorting them in ascending order. We then construct $\hat{Q}^{\mathsf{R}}(\tau|\bx)$ by applying linear interpolation similar to \eqref{eq:lin-interpo} to the rearranged sequence $\{(\tau_j,\tilde{Q}^{\mathsf{R}}(\tau_j|\bx)):j=1,\ldots,m-1\}$:
 
 In the online stage, we use $\hat{Y}^{\mathsf{R}}(x)\coloneqq \hat{Q}^{\mathsf{R}}(U|\bx)$ as the generative metamodel. %

 The rearrangement step adds minimal computational overhead to QRGMM. 
 As demonstrated by the above section, quantile crossing rarely occurs with $m=O(\sqrt{n})$, limiting how frequent rearrangement is needed. When invoked, this step is simply a sorting operation implementable via quicksort with average time complexity $O(m\log(m))$ \citep[Section~7.4]{Corman22intro}.
 
 Numerical experiments presented in Section \ref{e-sec:QRGMM-R-numerical} of the e-companion show that QRGMM with rearrangement (QRGMM-R) achieves slightly better data quality than the original QRGMM while maintaining computational efficiency. 
 The method ensures that we can effectively eliminate quantile crossing when it occurs without impacting online generation speed.%

 The rearrangement step not only adds minimal computational overhead but also preserves the asymptotic properties of QRGMM with linear quantile regression models established in Sections \ref{sec:asy} under additional mild conditions.
 The proofs are provided in Sections \ref{e-sec:convergence} of the e-companion.
 
 \begin{theorem}\label{Thm03} 
 	Under  assumptions specified in Section~\ref{e-sec:convergence}, 
 	the following results hold:
 	\begin{enumerate}[label=(2.\alph*), left=0pt]
 		\item For any $\bx^*\in \mathcal{X}$, $\hat{Y}^{\mathsf{R}}(\bx^*) \Rightarrow Y(\bx^*)$ as $n,m\to\infty$ . \label{thm:weakconvergence_a_R}	
 		\item For any $\bx^*\in \mathcal{X}$,
 		\[
 		\sup_{y\in\left(-\infty,\infty\right)}\left|F_{\hat{Y}^{\mathsf{R}}}(y|\bx^*)-F_Y(y|\bx^*)\right| \to 0,
 		\]
 		as $n,m\to\infty$, 
 		where $F_{\hat{Y}^{\mathsf{R}}}(y|\bx^*)$ is the conditional CDF of $\hat{Y}^{\mathsf{R}}(\bx^*)$ given $\bx^*$. \label{thm:weakconvergence_b_R}	
 	\end{enumerate} 
 \end{theorem}

In summary, combining the above insights, we recommend setting $m = O(\sqrt{n})$ for QRGMM. Numerical experiments in Section \ref{sec:choice_m} further support this choice, indicating that the approximation error, measured via the Kolmogorov-Smirnov statistic, stabilizes when $m$ exceeds approximately $\sqrt{n}$. This recommendation balances statistical accuracy and computational efficiency while addressing the trade-off associated with quantile crossing frequency, making it suitable for practical applications.

\section{Extensions} \label{sec:extension}

While the QRGMM framework is theoretically developed for one-dimensional outputs and linear quantile regression models, it extends to multidimensional outputs and other quantile regression models, as shown in this section. The theoretical analysis of these extensions, being fundamentally different from that of linear models, lies beyond this paper's scope. We evaluate their performance via numerical experiments on a complex multidimensional bank queueing simulation system in Section \ref{sec:bank-sym} in the e-companion.

\subsection{Multidimensional Outputs}

Extending the QRGMM framework to multidimensional outputs presents fundamental challenges because  the notion of ``joint quantiles'' for random vectors is not uniquely defined.
While various approaches for multi-output quantile regression exist in the literature, each offers distinct advantages and limitations, as discussed in \cite{CamehlFokGruber25} and references therein.

Rather than adopting multi-output quantile regression models that would fundamentally alter the QRGMM framework, we apply the \emph{conditional distribution method} \citep[page~555]{Devroye86}, which decomposes multivariate distribution generation into a sequence of univariate distribution generations, thereby preserving single-output quantile regression models. 

For a $d$-dimensional simulator output $\bY = (Y_1, \ldots, Y_d) \in \mathbb R^d$, generation of the conditional distribution of $\bY$ given $\bx$ proceeds as follows: generate $Y_1 = y_1$ from its conditional distribution given $\bx$, then sequentially generate $Y_l = y_l$ from its conditional distribution given $(\bx, \by_{[l-1]})$ for $l=2,\ldots,d$, where $\by_{[l-1]} = (y_1, \ldots, y_{l-1})$.  
Each conditional distribution generation step applies the inverse transform method, generating $F_{Y_l}^{-1}(U|\bx, \by_{[l-1]})$, where $F_{Y_l}^{-1}(\tau|\bx, \by_{[l-1]})$ denotes the conditional quantile function of $Y_l$ given $(\bx, \by_{[l-1]})$. 
The conditional quantile functions can be estimated using single-output quantile regression models in the offline stage. 
Algorithm \ref{alg:Mul-QRGMM} in the e-companion details this extension. %

\subsection{Nonlinear Quantile Regression Models}

One can address the potential nonlinearity of the conditional quantile function $F_Y^{-1}(\tau|\bx)$ in the covariates $\bx$ by incorporating basis functions in the linear quantile regression model. However, identifying an appropriate basis function set for a specific application remains challenging.

Neural network models have demonstrated superior performance in supervised learning tasks for predicting conditional expectations with highly nonlinear relationships \citep{goodfellow16deep}. These models adapt readily to conditional quantile prediction, though the non-differentiability of the pinball loss function complicates neural network training for quantile regression. \cite{cannon2011quantile} proposed using the Huber loss function to approximate the pinball loss function smoothly, enabling efficient gradient-based optimization. 

Section \ref{sec:bank-sym} of the e-companion shows that neural network quantile regression models can be effectively integrated with multi-output QRGMM, using a high-dimensional nonlinear bank queueing simulator with ten-dimensional covariates and five-dimensional outputs as a testbed. The model achieves strong empirical performance, demonstrating the scalability of the QRGMM framework.

\section{Numerical Experiments}\label{sec:num}

In this section, we conduct numerical experiments to evaluate QRGMM's performance, validate its asymptotic properties, and demonstrate its practical utility.
We compare QRGMM with several popular conditional generative baselines, including the conditional Wasserstein GAN with gradient penalty (CWGAN) \citep{AtheyImbensMetzgerMunro24}, denoising diffusion implicit models (Diffusion) \citep{song2021denoising}, and rectified flow (RectFlow) \citep{liu2023flow}. Our results demonstrate that QRGMM achieves superior performance and robustness in the task of simulation generative metamodeling.

We first evaluate all models using artificial test problems with known underlying conditional distributions, while also validating QRGMM's convergence properties and providing empirical insights about the choice of $m$.
Then, we assess all models for a simulator of esophageal cancer, demonstrating how generative metamodels can expedite simulation and enable real-time decision-making.

All experiments are conducted on a laptop equipped with an Intel 12th Gen CPU (14 cores, 20 threads, 2.7GHz), 16 GB RAM, and an NVIDIA GeForce RTX 3060 Laptop GPU (6 GB). For generative metamodels, both distributional accuracy and computational efficiency are crucial. Our results show that QRGMM achieves clear advantages over the competing methods on both criteria. The source code for reproducing all experiments, together with an easy-to-use codebase of QRGMM, is available at \url{https://github.com/Keaikai/Generative-Metamodeling}.

\subsection{Artificial Test Problems}\label{sec:synthetic}

Consider two test problems where both the conditional distribution $Y|\bx$ and the marginal distribution of $\bx$ are known.
Assume $x_1 \sim \mathsf{Unif}(0,10)$, $x_2\sim \mathsf{Unif}(-5,5)$, and $x_3 \sim \mathsf{Unif}(0,5)$ are independent.  

\begin{itemize}
    \item 
\textit{Test Problem~1}: $\bx=(1,x_1,x_2,x_3) \in \mathbb{R}^4$ and $Y|\bx \sim \mathcal{N}(\mu(\bx), \sigma^2(\bx))$, where 
$\mu(\bx)=5+x_1+2x_2+0.5x_3$ and $\sigma(\bx)=1+0.1x_1+0.2x_2+0.05x_3$.
The conditional quantile function satisfies Assumption~\ref{ass:cons1}:
\[
    F^{-1}_Y(\tau|\bx) =  \mu(\bx)+\sigma(\bx)z_{\tau}
	=  5+z_{\tau}+(1+0.1z_{\tau})x_1+(2+0.2z_{\tau})x_2+(0.5+0.05z_{\tau})x_3,
\]
where $z_{\tau}$ is the $\tau$-quantile of the standard normal distribution.
For this problem, we use linear quantile regression in QRGMM: $Q(\tau|\bx) = \bbeta(\tau)^\intercal \bx$. 
\item 
\textit{Test Problem~2}: $\bx=(1,x_1,x_2)\in\mathbb{R}^3$ and $Y|\bx$ follows a Laplace distribution with the location parameter $l(\bx)$ and scale parameter $s(\bx)$, where $l(\bx)=0.05x_1x_2$ and $s(\bx)=5\sin^2(x_1+x_2) +5$.
The conditional quantile function does NOT satisfy Assumption~\ref{ass:cons1}:
\[
    F^{-1}_Y(\tau|\bx)= l(\bx)+s(\bx)\zeta_{\tau}= 0.05x_1x_2+\left[5\sin^2(x_1+x_2) +5\right]\zeta_{\tau},
\]
where $\zeta_{\tau}$ is the $\tau$-quantile of the $\mathrm{Laplace}(0,1)$ distribution. 
For this problem, we use linear quantile regression with third-order polynomial basis functions in QRGMM: 
$Q(\tau|\bx)=\bbeta(\tau)^\intercal\mathbf{b(x)}$, where $\mathbf{b(x)}=(1,x_1,x_2,x_1^2,x_1x_2,x_2^2,x_1^3,x_1^2x_2,x_1x_2^2,x_2^3)$.
This test problem is intentionally constructed to violate Assumption~\ref{ass:cons1}, allowing us to examine the robustness of QRGMM under model misspecification.
\end{itemize}

\subsubsection{Performance Comparisons.}\label{subsec:syn_compare} 
\begin{figure}[t]
	\FIGURE{
		$\begin{array}{c}\\
			\includegraphics[width=0.8\textwidth,height=0.35\linewidth]{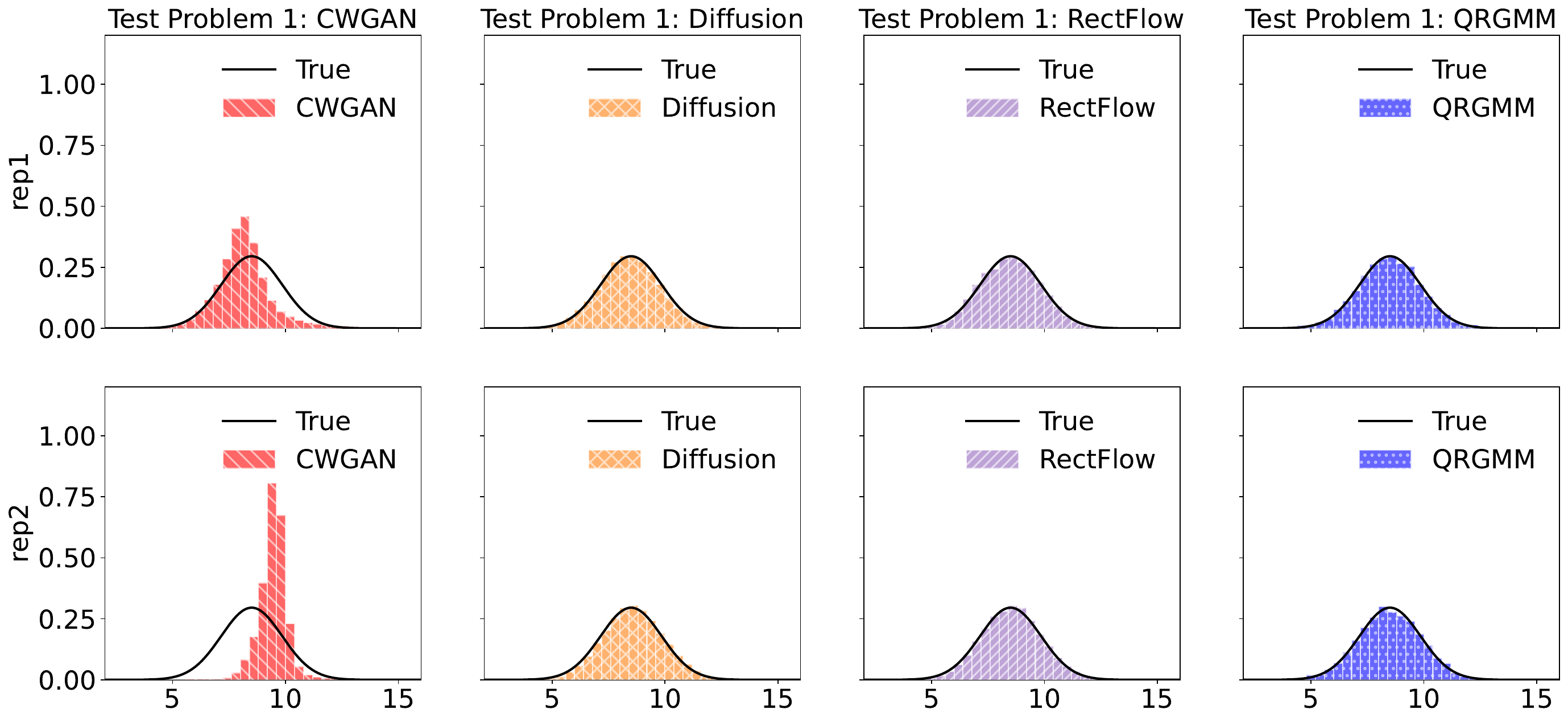} \\
			\includegraphics[width=0.8\textwidth,height=0.35\linewidth]{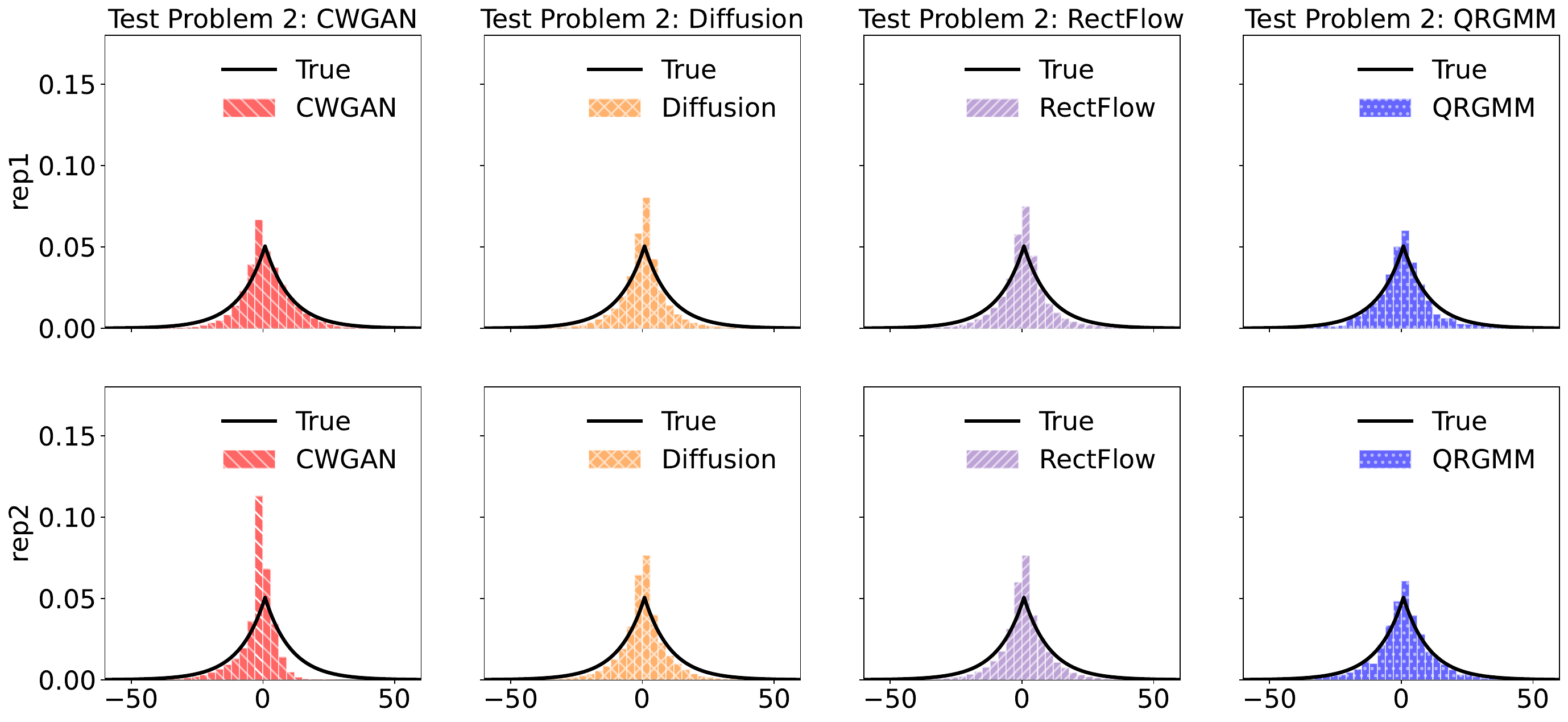}
		\end{array}$
	}
	{Generated Samples Approximating Conditional Distribution $Y$ Given $\bx^*$. 
		\label{fig_LQR_vs_hist01}} 
	{Test Problem~1: $\bx^*= (1,4,-1,3)$; Test Problem~2: $\bx^*= (1,4,4)$. ``rep1'' and ``rep2'' refer to two replications. %
	}
\end{figure}
For each test problem, we generate a training dataset $\{(\bx_i,y_i)\}_{i=1}^n$ of size $n=10^4$ to train all models.
For evaluation, we generate $K=10^5$ observations from each trained model at a new covariate vector $\bx^*$ and compare them with the known true conditional distribution.
We set QRGMM's grid size as $m=300$, adopt the CWGAN hyperparameters based on \citet{AtheyImbensMetzgerMunro24}, and use the Diffusion and RectFlow implementation with the hyperparameters in \citet{liang2024generative}.
We conduct multiple independent replications of this experiment.

Figure \ref{fig_LQR_vs_hist01} compares two replications of each method. QRGMM, Diffusion, and RectFlow all generate samples that align closely with the true distributions in both scenarios. In contrast, CWGAN exhibits pronounced instability across replications: despite using identical hyperparameter settings, it performs reasonably well in some runs but fails badly in others. This behavior is characteristic of the mode collapse phenomenon, in which generated samples concentrate in limited regions of the distribution, and is largely attributable to the intrinsic instability of adversarial training, which involves solving a tricky saddle-point optimization problem \citep{saxena2021generative}.

While Figure \ref{fig_LQR_vs_hist01} demonstrates performances for specific $\bx^*$ and replications, we also evaluate the methods from an ``unconditional'' perspective using the marginal distribution of $Y$.
We generate a test dataset $\{(\bx^{\prime}_k,y^{\prime}_k)\}_{k=1}^K$ from the true distribution, then use each trained metamodel to generate one approximate sample $\hat{y}^{\prime}_k$ for each $\bx^{\prime}_k$.
We compare the resulting approximate sample $\{\hat{y}^{\prime}_k\}_{k=1}^K$ and the true sample $\{y^{\prime}_k\}_{k=1}^K$ of the (unconditional) distribution of $Y$. 
We calculate the mean and standard deviation of the two samples, along with two statistical distance measures between them: the Wasserstein distance \citep{arjovsky2017wasserstein} and Kolmogorov--Smirnov (KS) statistic \citep{massey1951kolmogorov}. 
Over 100 replications, we compute the mean and standard deviation of these summary statistics, with results reported in Table~\ref{table:sync} and Figure~\ref{fig_syn_WDKS}.

\begin{table}[t]
	\TABLE{Generated Samples Approximating Distribution of $Y$: Mean and Standard Deviation. \label{table:sync}}
	{\begin{tabular}{ccccccccccccccccc}
			\toprule 
			\multirow{3}{*}{} && \multicolumn{7}{c}{Sample Mean} && \multicolumn{7}{c}{Sample SD}\\
			\cmidrule(l){3-9} 
			\cmidrule(l){11-17} 
			&& \multicolumn{3}{c}{Test Problem~1}    && \multicolumn{3}{c}{Test Problem~2} && \multicolumn{3}{c}{Test Problem~1} 
			&& \multicolumn{3}{c}{Test Problem~2} \\ 
			\cmidrule(l){3-5} \cmidrule(l){7-9} \cmidrule(l){11-13} \cmidrule(l){15-17} 
			&& {Mean}       && {SD}          && {Mean}          && {SD} 
			&& {Mean}       && {SD}          && {Mean}          && {SD}         \\ 
			\cmidrule(l){1-1} \cmidrule(l){3-3} \cmidrule(l){5-5} \cmidrule(l){7-7} \cmidrule(l){9-9} \cmidrule(l){11-11} \cmidrule(l){13-13} \cmidrule(l){15-15} \cmidrule(l){17-17} 
			
			Truth 
			&& $11.2469$ && $0.0633$ && $-0.0007$ && $0.1056$
			&& $6.7322$  && $0.0410$ && $10.9196$ && $0.1165$ \\
			
			QRGMM 
			&& $11.2491$ && $0.0652$ && $-0.0050$ && $0.1209$
			&& $6.7290$  && $0.0442$ && $10.4025$ && $0.1576$ \\
			
			CWGAN 
			&& $11.2424$ && $1.6882$ && $\phantom{-}0.0884$ && $2.3263$
			&& $5.9863$  && $1.1323$ && $\phantom{1}9.7939$ && $2.0161$ \\
			
			Diffusion
			&& $11.2448$ && $0.1051$ && $\phantom{-}0.0153$ && $0.3431$
			&& $6.7263$  && $0.0762$ && $9.5797$ && $0.6252$ \\
			
			RectFlow
			&& $11.2556$ && $0.0889$ && $-0.0081$ && $0.1885$
			&& $6.7178$  && $0.0691$ && $9.4219$ && $0.3306$ \\
			
			\bottomrule
		\end{tabular}
	}
	{``Sample Mean'' and ``Sample SD'' are the mean and standard deviation of $\{\hat{y}^{\prime}_k\}_{k=1}^K$ or $\{y^{\prime}_k\}_{k=1}^K$, respectively. The results are calculated based on 100 replications.}	
\end{table}

\begin{figure}
\FIGURE{
$\begin{array}{c}
\includegraphics[width=0.6\textwidth,height=0.25\linewidth]{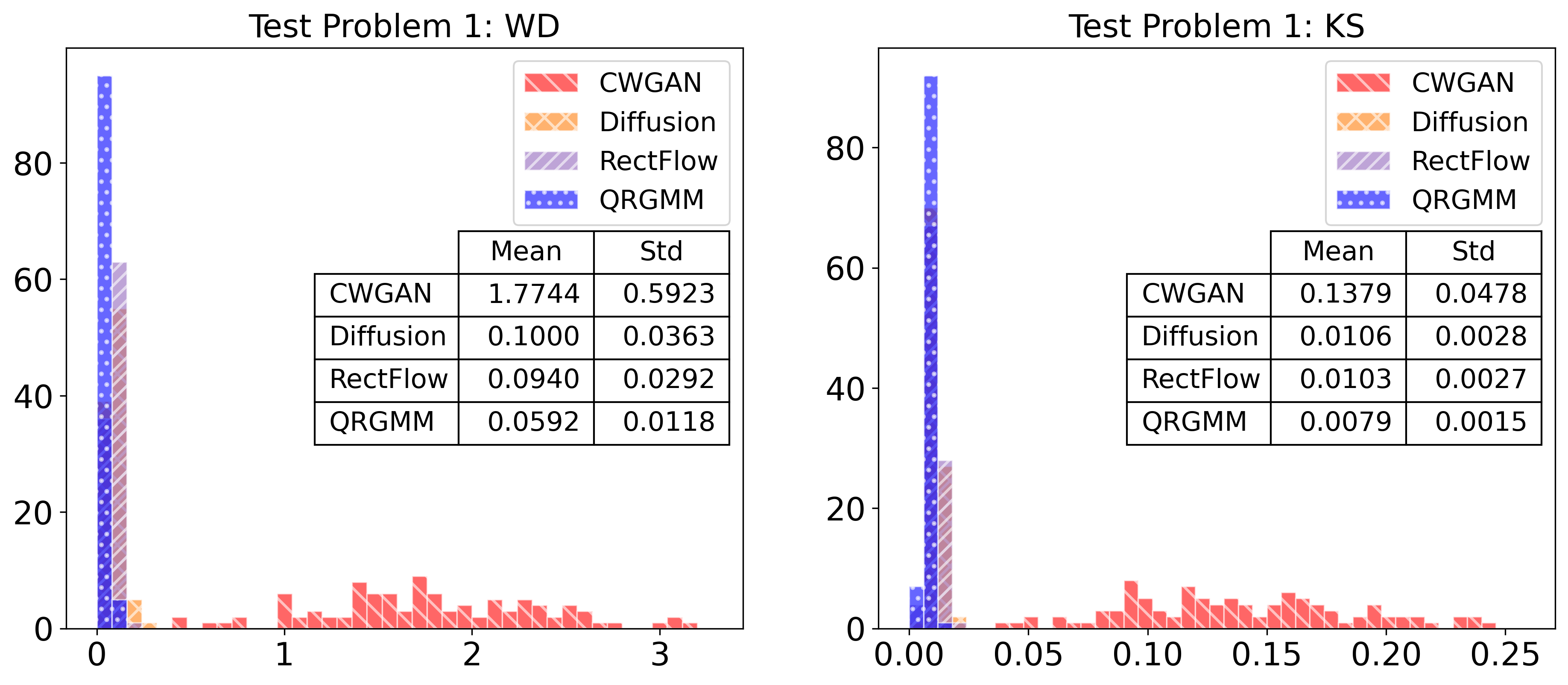}\\ 
\includegraphics[width=0.6\textwidth,height=0.25\linewidth]{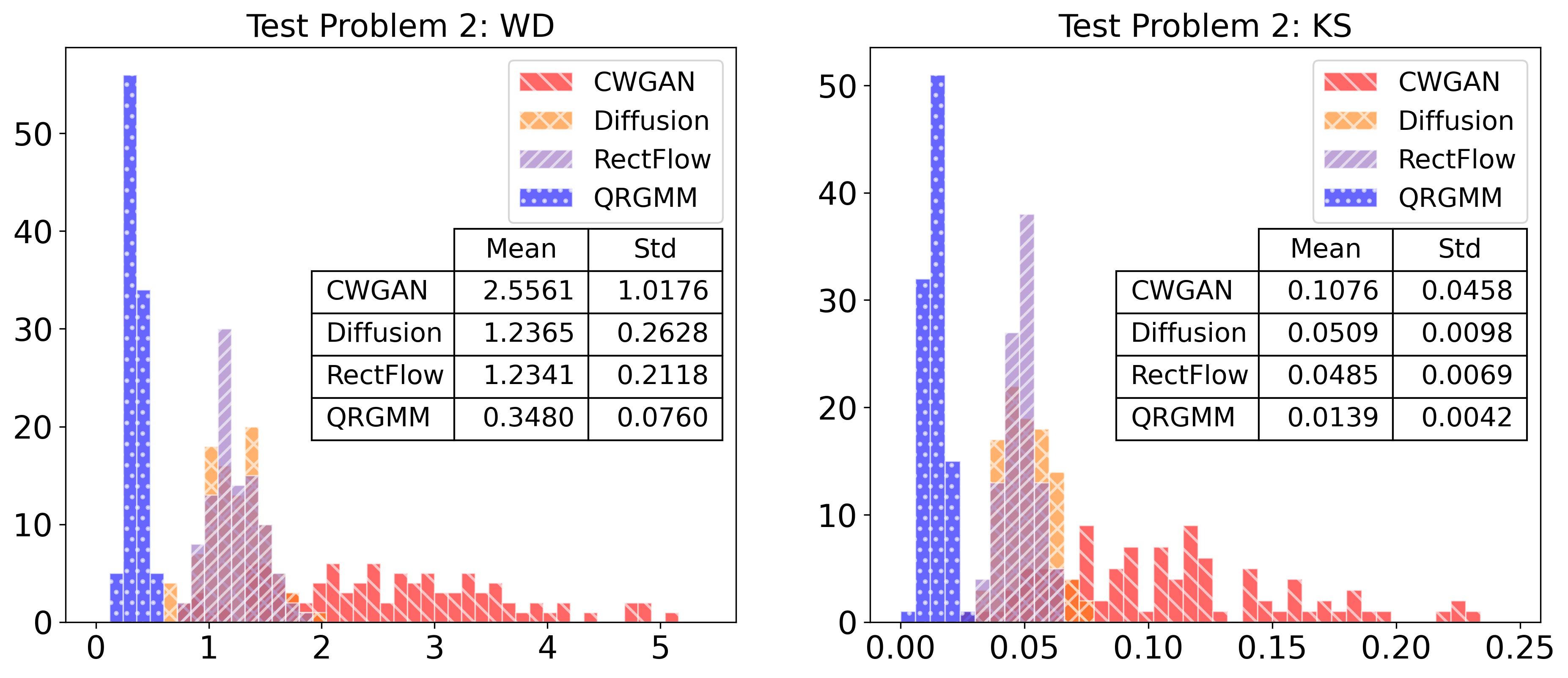}
 \end{array}$
 }
{Generated Samples Approximating $Y$ Distribution: Wasserstein Distance (WD) and KS Statistic (KS).  \label{fig_syn_WDKS} }
{WD and KS are calculated between $\{\hat{y}^{\prime}_k\}_{k=1}^K$ and $\{y^{\prime}_k\}_{k=1}^K$ over 100 replications.}
\end{figure}

Table~\ref{table:sync} shows that QRGMM-generated samples more accurately match the true mean and standard deviation compared to generated samples from other models.
QRGMM also exhibits lower variability across replications in both test problems, as evidenced by smaller standard deviations of sample means and standard deviations.
Figure~\ref{fig_syn_WDKS} further confirms QRGMM's superiority through consistently smaller Wasserstein distances and KS statistics, with notably lower variation across replications.

\subsubsection{Convergence in Distribution.}\label{subsec:converg_dist}
\begin{figure}[t]
	\FIGURE{
		\hspace{0.7em}
		\includegraphics[width=0.575\linewidth]{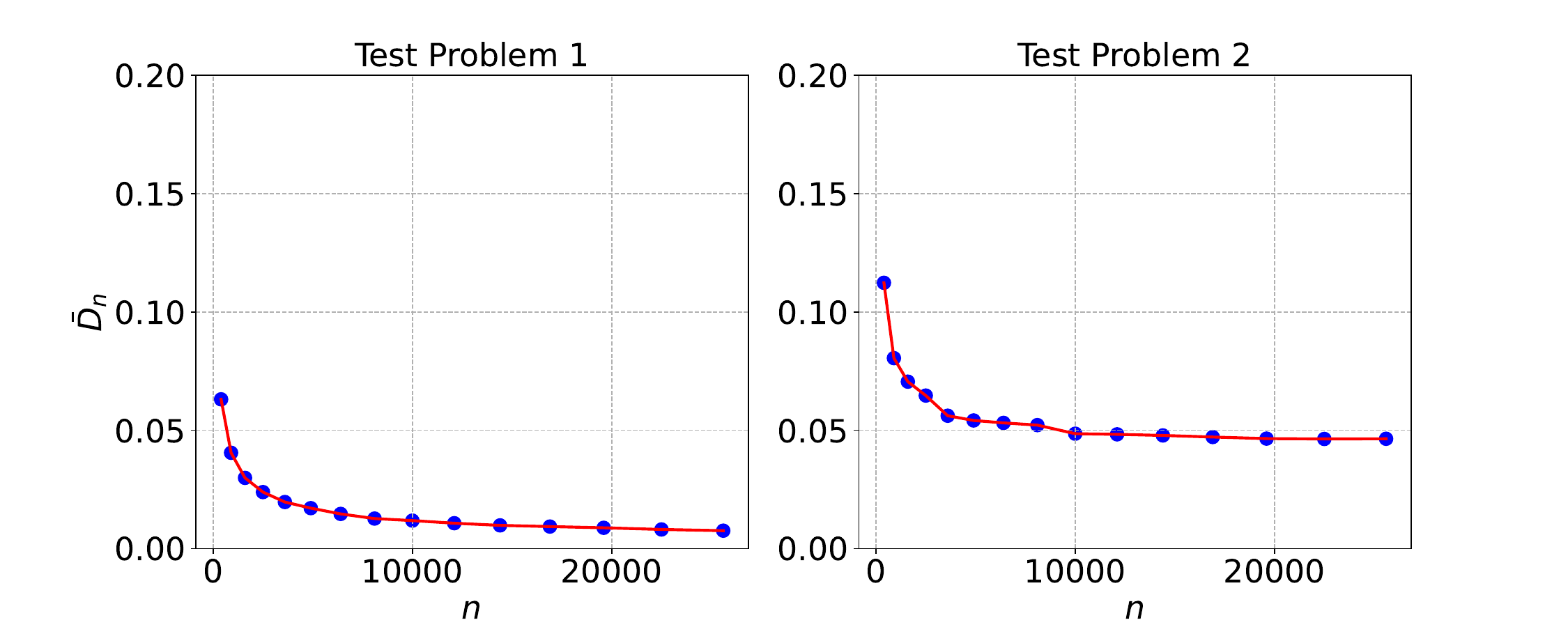}
		\hspace{-3.65em}
		\includegraphics[width=0.575\linewidth]{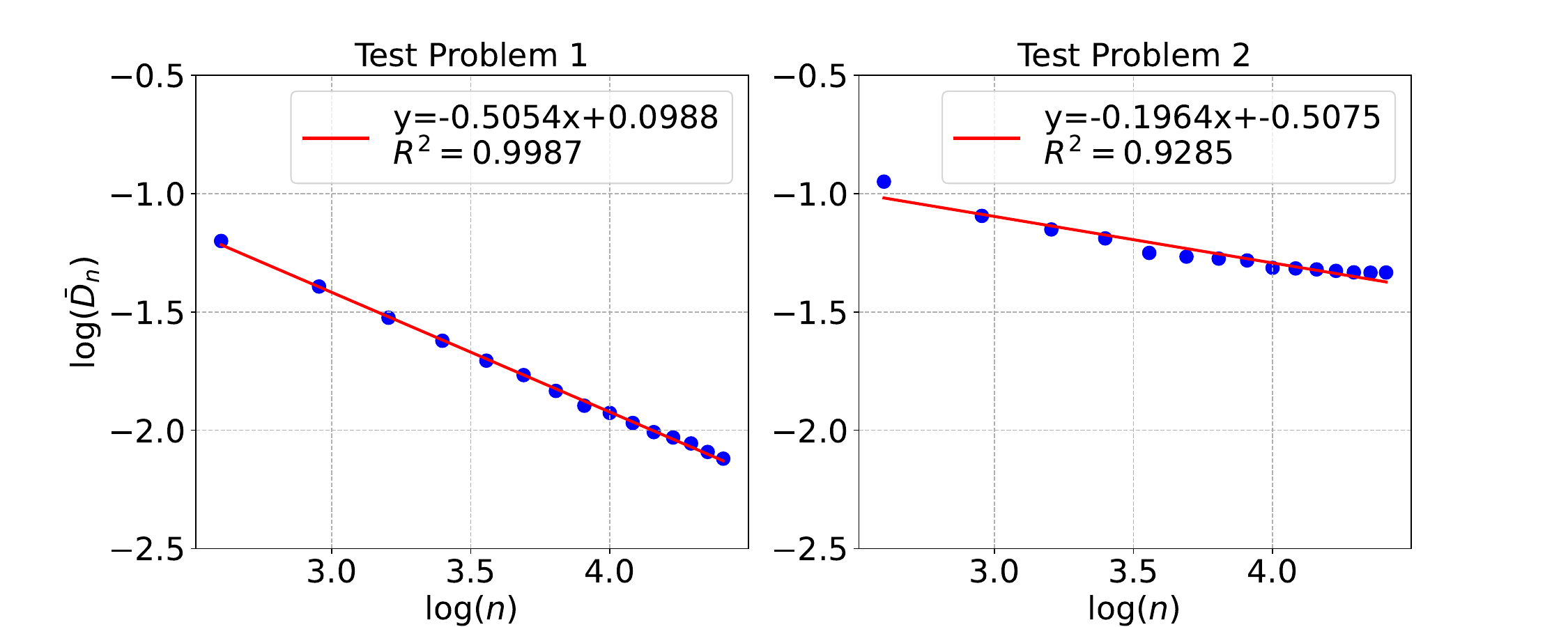}
	}
	{Convergence of KS Statistic as $n$ Increases. \label{fig:logdvslogn_scenario12}}
	{$\bar{D}_n \approx \sup_y |F_{\hat{Y}}(y|\bx^*)-F_Y(y|\bx^*)| $. Test Problem~1: $\bx^*=(1,6,1,2)$; Test Problem 2: $\bx^*=(1,7,2)$.   }
\end{figure}

We demonstrate the uniform convergence of the conditional CDF of $\hat{Y}(\bx^*)$ given $\bx^*$ as stated in Theorem~\ref{thm:weakconvergence_b}:
$\sup_y |F_{\hat{Y}}(y|\bx^*)-F_Y(y|\bx^*)| \to 0$ as $n\to\infty$.
This supremum defines the KS distance between distributions.
Since $F_{\hat{Y}}(y|\bx^*)$ lacks an explicit expression, we approximate it using the empirical distribution of a large generated sample.

For each training dataset of size $n$, we train QRGMM with $m=\sqrt{n}$ and generate $K=10^5$ samples using the trained metamodel $\hat{Y}(\bx^*)$. 
The KS statistic ($D_n$) is then computed between the sample's empirical distribution and the true distribution of $Y(\bx^*)$.
We repeat this process 100 times for each $n=m^2$ with $m=20, 30, \ldots, 160$ and compute $\bar{D}_n$, the average KS statistic across replications.

Figure~\ref{fig:logdvslogn_scenario12} displays the relationship between $\bar{D}_n$ and $n$. For Test Problem~1 with $\bx^*=(1,6,1,2)$, $\bar{D}_n$ converges to zero, confirming QRGMM's convergence in distribution to the conditional distribution of $Y$ given $\bx^*$. And the slope in the log--log plot is close to the $n^{-1/2}$ rate value discussed in Section~\ref{m_choice_rate}, with an $R^2$ close to one. In contrast, for Test Problem~2 with $\bx^*=(1,7,2)$, where Assumption~\ref{ass:cons1} is not satisfied, $\bar{D}_n$ converges to a positive value and the log--log slope deviates from this rate due to model misspecification induced by the polynomial approximation when Assumption~\ref{ass:cons1} does not hold. By varying the training sample size $n$, this experiment also illustrates how the method’s finite-sample performance depends on $n$. Notably, we observe that in both test problems, the model achieves satisfactory performance once the sample size exceeds $n=2500$, demonstrating that the method is data-efficient and performs well in practice even with moderate sample sizes.

\subsubsection{Choice of $m$.}\label{sec:choice_m}

In this subsection, we explore how to choose the parameter $m$ in practice and aim to provide some other empirical insights. To this end, we conduct experiments using Test Problem~1 with a training dataset of size $n=10^4$. For each value of $m=10^1, 10^{1.25}, \ldots, 10^{3.75}, 10^4$, we train QRGMM and generate $K=10^5$ samples from the fitted model $\hat{Y}(\bx^*)$. We then compute the average KS statistic ($D_m$) between the empirical distribution of $\hat{Y}(\bx^*)$ and the true distribution of $Y(\bx^*)$ over 100 replications. As shown in Figure~\ref{fig:dvsm}, $\bar{D}_m$ decreases as $m$ increases, but the improvement saturates when $m$ exceeds approximately $\sqrt{n}=100$. This suggests that choosing $m$ proportional to $\sqrt{n}$ achieves a good balance between estimation accuracy and computational cost in practice.

\begin{figure}[t]
	\FIGURE{
		\includegraphics[width=0.6\linewidth]{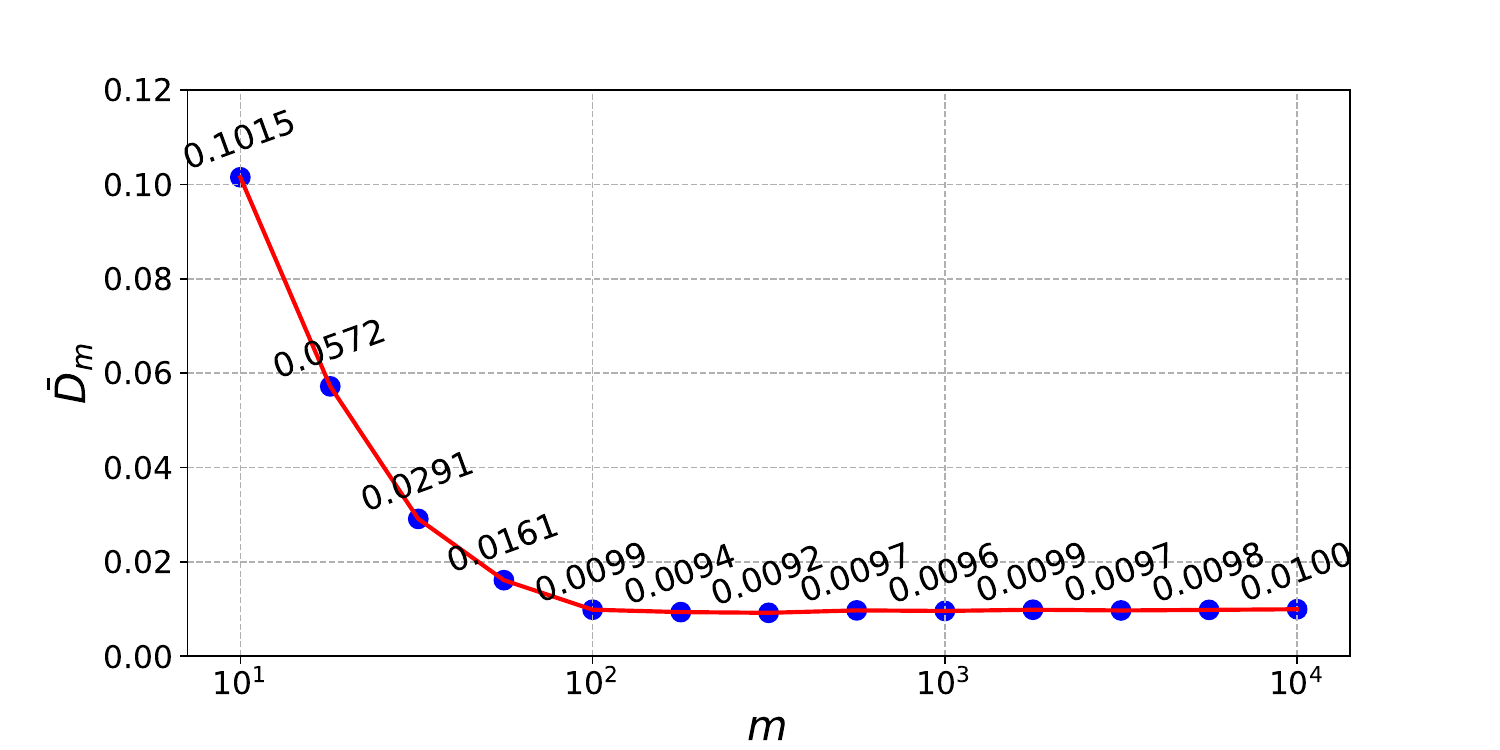} 
	}
	{Effect of $m$ on KS Statistic.
		\label{fig:dvsm} }
	{Test Problem 1 with $\bx^*=(1,6,1,2)$. The training dataset has size $n=10^4$. }
\end{figure}

\subsection{Simulation for Esophageal Cancer Treatments}\label{sec:practical}

We apply QRGMM to an esophageal cancer simulation model, comparing its performance with other models. Developed by \cite{hur2004cost} and \cite{choi2014statins}, this  model evaluates aspirin and statin chemoprevention strategies for \emph{esophageal adenocarcinoma} (EAC) and \emph{Barrett's esophagus} (BE) management. EAC, a predominant esophageal cancer subtype whose incidence has increased dramatically in recent decades \citep{choi2014statins}, develops from BE as a precursor condition. While aspirin and statin effectively reduce BE-to-EAC progression \citep{kastelein2011nonsteroidal}, patient outcomes depend on factors including age, weight, lifestyle habits, and individual drug responses.
The model simulates quality-adjusted life years (QALYs) by tracking health state transitions until death for BE patients under various treatment regimens. It serves as a benchmark for evaluating methods that solve \emph{ranking and selection with covariates} (R\&S-C) problems \citep{shen2021ranking,li2024efficient}.
The code is available at \url{https://simopt.github.io/ECSim}.

The model has four inputs $\bx = (x_1, x_2, x_3, x_4)$, where $x_1 \in [55,80]$ is the treatment start age \citep{naef1972conservative}, $x_2 \in [0,0.1]$ is the annual BE to EAC progression rate \citep{hur2004cost}, $x_3 \in [0,1]$ is aspirin's progression reduction effect, and $x_4 \in [0,1]$ is statin's progression reduction effect. 
The model outputs quality-adjusted life years (QALYs) for two treatments: $Y_1(\bx)$ for aspirin and $Y_2(\bx)$ for statin. Given their independence, the conditional distributions given $\bx$ can be estimated separately.

\subsubsection{Treatment Effects.}\label{treatment_effects}

\begin{table}[t]
	\TABLE{Conditional Average Treatment Effects and Treatment Ranking. \label{table_EC_vs_utility} }
	{\begin{tabular}{ccccccccccc}
			\toprule 
			\multirow{2}{*}{} 
			&& \multicolumn{3}{c}{$\bar{Y}_1(\bx^*)$}     && \multicolumn{3}{c}{$\bar{Y}_2(\bx^*)$} && \multirow{2}{*}{Percentage of Correct Ranking}         \\ 
			\cmidrule(l){3-5} \cmidrule(l){7-9} 
			&& {Mean}       && {SD}          && {Mean}          && {SD} \\ 
			\cmidrule(l){1-1} \cmidrule(l){3-3} \cmidrule(l){5-5} \cmidrule(l){7-7} \cmidrule(l){9-9} \cmidrule(l){11-11}
			
			Truth              
			&& $13.4445$ && $0.0235$ && $14.0217$ && $0.0243$ && $100\%$ \\ 
			
			QRGMM             
			&& $13.4703$ && $0.2041$ && $14.0419$ && $0.1811$ && $100\%$ \\ 
			
			LinearReg          
			&& $13.4703$ && $0.2047$ && $14.0418$ && $0.1928$ && $\phantom{1}98\%$ \\ 
			
			CWGAN             
			&& $13.4548$ && $2.2922$ && $14.3274$ && $2.1655$ && $\phantom{1}61\%$ \\ 
			
			Diffusion
			&& $13.4024$ && $0.3883$ && $13.9783$ && $0.4240$ && $\phantom{1}82\%$ \\
			
			RectFlow
			&& $13.3027$ && $0.8664$ && $14.1208$ && $0.8363$ && $\phantom{1}75\%$ \\
			
			\bottomrule
		\end{tabular}
	}
	{The correct ranking for $\bx^*$ is $\mathbb{E}[Y_2(\bx^*)] > \mathbb{E}[Y_1(\bx^*)]$. The results are calculated based on 100 replications. }	
\end{table}

Consider estimating the conditional average treatment effect $\mathbb{E}[Y_l(\bx^*)]$ (i.e., the expected QALYs) for treatments $l=1,2$ given a patient's covariate $\bx^*$ and ranking these treatments. 
We generate a training dataset $\{(\bx_i, y_{1i}, y_{2i})\}_{i=1}^n$ with $n=10^4$, where $\bx_i$ follows the uniform distribution over its domain, and $y_{l,i}$ is a sample of $Y_l(\bx_i)$. 
For each output dimension~$l$, we train both QRGMM with $m=300$ and the quantile regression model $Q(\tau|\bx) = \bbeta(\tau)^\intercal \bb(\bx)$ on $\{(\bx_i, y_{l,i})\}_{i=1}^n$, where $\bb(\bx)$ represents second-order polynomial basis functions, and all other models keep the same hyperparameter settings taken from \citet{AtheyImbensMetzgerMunro24} and \citet{liang2024generative}.

For a patient with covariate vector $\bx^*=(1,70,0.05,0.2,0.8)$, we generate $K=10^5$ approximate samples of $Y_l(\bx^*)$ using each trained metamodel and compute the sample mean $\bar{Y}_l(\bx^*)$ for each treatment $l=1,2$. 
We rank the treatments based on these sample means. 
We also compare these generative metamodels with the linear regression (LR) metamodel from \cite{shen2021ranking}, which estimates expected treatment effects as a function of covariates. 
For fair comparison, we use identical polynomial basis functions for LR and QRGMM.
We approximate the true values of $\mathbb{E}[Y_1(\bx^*)]$ and $\mathbb{E}[Y_2(\bx^*)]$ using $n=10^5$ samples from the simulator, although this requires much more computational time.

We repeat these experiments 100 times and calculate the means and standard deviations of $\bar{Y}_1(\bx^*)$ and $\bar{Y}_2(\bx^*)$ along with the percentage of correct ranking. From extensive simulator runs, we know treatment 2 is superior to treatment 1 for this covariate $\bx^*$. %
The results are shown in Table \ref{table_EC_vs_utility}. QRGMM generates data that effectively estimates and compares means of $Y_1(\bx^*)$ and $Y_2(\bx^*)$, performing similarly to real data and the LR metamodel while significantly outperforming other models. 
LR's strong performance reflects the simulator's smooth, approximately linear response surface noted by \cite{shen2021ranking}. 
Although we focus here on estimation of means where traditional metamodeling methods excel, 
QRGMM shows competitive performance even in these scenarios favorable to LR.

\begin{table}[t]
	\TABLE{Computational Time for Training Generative Metamodels and Generating $K=10^5$ Samples (in Seconds). \label{table_EC_vs_time}}
	{\begin{tabular}{lcccccc}
			\toprule
			& \multicolumn{2}{c}{Test Problem~1} 
			& \multicolumn{1}{c}{Test Problem~2} 
			& \multicolumn{2}{c}{Esophageal Cancer Simulator} \\
			\cmidrule(lr){2-3}\cmidrule(lr){4-4}\cmidrule(lr){5-6}
			& Train Time & Gen Time & Gen Time 
			& $Y_1(\bx^*)$ Gen Time & $Y_2(\bx^*)$ Gen Time \\
			\midrule
			
			QRGMM     
			& $2.5561\,(\pm0.3602)$ & $0.0032\,(\pm0.0004)$ & $0.0033\,(\pm0.0006)$
			& $0.0017\,(\pm0.0006)$ & $0.0017\,(\pm0.0006)$ \\
			
			CWGAN     
			& $96.0983\,(\pm1.7349)$ & $0.2018\,(\pm0.0264)$ & $0.2022\,(\pm0.0270)$
			& $0.2287\,(\pm0.0199)$ & $0.2092\,(\pm0.0300)$ \\
			
			Diffusion 
			& $34.6643\,(\pm2.9444)$ & $0.8229\,(\pm0.0181)$ & $0.8207\,(\pm0.0094)$
			& $0.8402\,(\pm0.0151)$ & $0.8384\,(\pm0.0140)$ \\
			
			RectFlow  
			& $33.9495\,(\pm2.8513)$ & $0.7646\,(\pm0.0128)$ & $0.7587\,(\pm0.0075)$
			& $0.7759\,(\pm0.0096)$ & $0.7726\,(\pm0.0095)$ \\
			
			\bottomrule
		\end{tabular}
	}
	{Each entry is reported as mean $(\pm$ standard deviation$)$ across 100 replications. 
		All times are measured in seconds. Gen Time is the computational time to generate $10^5$ conditional observations for a given $\bx^*$.}
\end{table}

Table~\ref{table_EC_vs_time} reports the computational time for training generative metamodels and generating $10^5$ conditional observations across three experiments. QRGMM trains quantile regressions over a grid of quantile levels, which can be naturally parallelized on CPUs as quantile regressions at different quantile levels are mutually independent; other models use GPUs for acceleration in both training and generating. All models have relatively short training times, with QRGMM being the fastest. Since this study emphasizes online generation speed, training time is reported only for Test Problem~1. Regarding generation, all models are capable of supporting real-time decision making. In particular, \emph{QRGMM requires less than $0.01$ seconds to generate $10^5$ conditional observations}, which is orders of magnitude faster than the other generative baselines, making it particularly attractive for time-critical applications. The faster generation times of QRGMM in the Esophageal Cancer experiment are mainly due to implementation. QRGMM is implemented in MATLAB for this case, benefiting from highly optimized matrix operations, while the synthetic experiments are run in Python. %

Lastly, following Section~\ref{subsec:syn_compare}, we compare all models using summary statistics of the unconditional distribution of $Y_l$ for $l=1,2$. 
For each test point $\bx_k'$ in dataset $\{(\bx_k', y_{l,k}')\}_{k=1}^K$, we generate one approximate sample $\hat{y}_{l,k}'$ using each trained metamodel. We compute means, standard deviations, Wasserstein distances, and KS statistics for both $\{\hat{y}_{l,k}'\}_{k=1}^K$ and $\{y_{l,k}'\}_{k=1}^K$. Over 100 repetitions, we calculate means and standard deviations of these summary statistics, with results in Table~\ref{table:EC_vs_MSD} and Figure \ref{fig_ECS_vs_hist_WDKS_Y12} in the e-companion. Consistent with Section~\ref{subsec:syn_compare}, QRGMM shows superior effectiveness and stability in approximating the conditional distribution of $Y(\bx^*)$ given $\bx^*$.

\subsubsection{Probability of Correct Selection.}

Selecting the optimal treatment among finite alternatives for each patient can be formulated as an R\&S-C problem, where policies are learned from offline simulations for online decisions. 
Time constraints typically prevent running simulations \emph{after observing patient covariates}. 
However, generative metamodels enable rapid sampling, allowing us to generate and compare numerous observations across treatments on the fly.

For each patient with covariates $\bx$, we generate $K=10^5$ observations per treatment using all models, and identify the optimal treatment by comparing sample means. 
We compare these generative metamodels against $\text{TS}^+$, an LR-based method proposed by \cite{shen2021ranking}, and the KN procedure of \cite{kim2001fully} using true simulator samples. 
While KN with the true simulator provides a theoretical benchmark, its computational intensity makes it impractical for real-time decision-making.

A key performance metric is the \emph{average probability of correct selection} ($\text{PCS}_{\text{E}}$), defined as:
\(
\text{PCS}_{\text{E}} \coloneqq \mathbb{P}\left(\mu_{l^*(\bx)}(\bx) - \mu_{\widehat{l^*}(\bx)}(\bx) < \delta \right),
\)
where $\mu_l(\bx) \coloneqq \mathbb{E}[Y_l(\bx)|\bx]$ denotes the conditional expected performance of treatment $l$,
$l^*(\bx)$ denotes the optimal treatment for covariates $\bx$, $\widehat{l^*}(\bx)$ denotes its estimate, and $\delta$ is the \emph{indifference-zone parameter}, representing the smallest meaningful difference to the decision-maker. We estimate  $\text{PCS}_{\text{E}}$ via 
\(
\widehat{\text{PCS}_{\text{E}}}=\frac{1}{R} \sum_{r=1}^R \frac{1}{T} \sum_{t=1}^T \mathbb{I}\left\{\mu_{l^*(\bx_{r,t})}(\bx_{r,t})-\mu_{\widehat{l^*_r}(\bx_{r,t})}(\bx_{r,t})<\delta\right\}, 
\)
where $R=100$ is the number of replications, $T=100$ is the number of patients, and $\delta=1/6$ QALYs. 
For each replication $r$, we generate covariates $\{\bx_{r,t}\}_{t=1}^T$ uniformly from their domain and determine the optimal treatment $l^*(\bx_{r,t})$ and its mean effect $\mu_{l^*(\bx_{r,t})}(\bx_{r,t})$ through brute-force simulation. 
All methods produce estimates $\widehat{l^*_r}(\bx_{r,t})$. 
We set the target $\text{PCS}_{\text{E}}$ to 95\% for $\text{TS}^+$ and KN.
Moreover, we use training size $n=10^5$ for generative metamodels, comparable to the sample size required by $\text{TS}^+$.

Table \ref{table_EC_RS} shows that QRGMM, Diffusion, and RectFlow achieve comparable performance to $\text{TS}^+$ and KN, all exceeding the 95\% target level, demonstrating their effectiveness for real-time decision-making. 
In contrast, CWGAN's lower performance indicates insufficient distributional accuracy for practical simulation-based decision-making.

\begin{table}[t]
	\TABLE{$\text{PCS}_{\text{E}}$ Achieved by Different Procedures. \label{table_EC_RS} }
	{\begin{tabular}{ccccccccccccc}
			\toprule 
			&& $\text{TS}^+$ && KN + Simulator && QRGMM && CWGAN && Diffusion && RectFlow \\ 
			\midrule 
			$\widehat{\text{PCS}_{\text{E}}}$ 
			&& $0.9946$ && $0.9952$ && $0.9858$ && $0.7480$ && $0.9830$ && $0.9716$ \\ 
			\bottomrule
		\end{tabular}
	}
	{}	
\end{table}

Unlike traditional R\&S-C procedures that require pre-specified summary statistics for comparing alternatives (e.g., $\text{TS}^+$ ranks by expected performance but cannot handle quantile criteria), generative metamodels allow flexible selection of evaluation metrics at decision time. 
In treatment selection, patients may evaluate options using criteria beyond expected effects. 
Generative metamodels accommodate this real-time selection of single or multiple comparison metrics.

\section{Conclusions}\label{sec:conclusions}

We introduce the concept of 
``generative metamodeling,'' which uses offline simulation to construct a ``fast simulator of the simulator'' for real-time decision-making. 
We propose QRGMM as a general framework independent of the specific quantile regression model. 
For linear models, we establish QRGMM's convergence in distribution. For practical implementation, we provide guidance on the choice of $m$. We further extend QRGMM to accommodate multidimensional outputs and neural network quantile regression models.

Our numerical results show that QRGMM, RectFlow, and Diffusion (denoising diffusion implicit models) can all be used to construct generative metamodels, whereas CWGAN fails to reliably capture output distributions. More importantly, in the one-dimensional and low-dimensional output settings common in stochastic simulation and operations research, QRGMM exhibits clear advantages in distributional accuracy, training efficiency, and sample generation speed, making it a natural choice for generative metamodeling in real-time decision-making, while more general deep generative models may constitute unnecessary overkill.

This initial exploration of generative metamodeling in stochastic simulation opens several research directions.
While our numerical experiments in the e-companion validate QRGMM extensions to multidimensional outputs and neural networks, their theoretical foundations require further study. 
Future work can also explore applications in real-world domains such as supply chain optimization, financial risk management, and healthcare systems.

\begingroup
\setlength{\bibsep}{0pt plus 0.3ex}
\bibliographystyle{informs2014} %
\bibliography{GenMetamodel.bib} %
\endgroup

\newpage

\ECSwitch

\EquationsNumberedBySection 
\ECHead{Supplementary Materials}

\section{Proof of Theorem \ref{thm:weakconvergence} (Convergence in Distribution)} 

\subsection{Proof of Lemma \ref{lemma:I1245}}

For $i=1,5$, 
\begin{eqnarray*}
	\mathbb{E}\left[ \left|h\left( \hat{Q}(U|\bx^*) \right) - h\left( F^{-1}_Y(U|\bx^*) \right)\right|\cdot \mathbb{I}\{U\in I_i\} \right] \le 2M_h\cdot\Pr(U\in I_i) = \frac{2M_h\cdot \epsilon}{10\max(M_h,1)} \leq \frac{\epsilon}{5}.
\end{eqnarray*}
For $i=2,4$, 
\begin{eqnarray*}
	\mathbb{E}\left[ \left|h\left( \hat{Q}(U|\bx^*) \right) - h\left( F^{-1}_Y(U|\bx^*) \right)\right|\cdot \mathbb{I}\{U\in I_i\} \right] \le 2M_h\cdot\Pr(U\in I_i) = \frac{2M_h}{m} \leq \frac{\epsilon}{5},
\end{eqnarray*}
provided that $m\ge {10\max(M_h,1)/\epsilon}$. \halmos

\subsection{Proof of Lemma \ref{lemma:I3} }

Note that 
\begin{equation}\nonumber %
	\left|h\left( \hat{Q}(U|\bx^*) \right) - h\left( F^{-1}_Y(U|\bx^*) \right)\right|\cdot \mathbb{I}\{U\in I_3\}\le L_h\cdot\sup_{\tau\in I_3} \left|\hat{Q}(\tau|\bx^*)-F^{-1}_Y(\tau|\bx^*)\right|.
\end{equation}
By Proposition \ref{pro:asym2},  as $n, m\to\infty$, 
\[
\sup_{\tau\in I_3}\left|\hat{Q}(\tau|\bx^*)-F^{-1}_Y(\tau|\bx^*)\right| \overset{\mathbb P}{\to} 0.
\]
Therefore,
\begin{equation}\label{pf:hp}
	\left|h\left( \hat{Q}(U|\bx^*) \right) - h\left( F^{-1}_Y(U|\bx^*) \right)\right|\cdot \mathbb{I}\{U\in I_3\}\overset{\mathbb P}{\to} 0.
\end{equation}
Moreover, because $h$ is bounded, 
\begin{equation}\label{pf:unint}
	\left|h\left( \hat{Q}(U|\bx^*) \right) - h\left( F^{-1}_Y(U|\bx^*) \right)\right|\cdot \mathbb{I}\{U\in I_3\}\le 2 M_h.
\end{equation}
It then follows from \eqref{pf:hp}, \eqref{pf:unint}, and the dominated convergence theorem that
\[
\mathbb{E}\left[ \left|h\left( \hat{Q}(U|\bx^*) \right) - h\left( F^{-1}_Y(U|\bx^*) \right)\right|\cdot \mathbb{I}\{U\in I_3\} \right]\to 0
\]
as $m,n\to\infty$. 
Therefore, for a given $\epsilon>0$, there exist sufficiently large $M$ and $N$, such that 
\[
\mathbb{E}\left[ \left|h\left( \hat{Q}(U|\bx^*) \right) - h\left( F^{-1}_Y(U|\bx^*) \right)\right|\cdot \mathbb{I}\{U\in I_3\} \right]\le{\frac{\epsilon}{5}},
\]
for all $m\geq M$, $n\geq N$.
\halmos

\subsection{Proof of Proposition \ref{pro:asym2}}
The notation $\tilde{\bbeta}_j$ is hereinafter employed to represent $\tilde{\bbeta}(\tau_j)$ for the sake of simplicity, while $\hat{\bbeta}(\tau)$ denotes QRGMM's estimation of $\bbeta(\tau)$ through interpolation for any given value of $\tau$. Notice that for any given subinterval $[\tau_{\mathsf{L}},\tau_{\mathsf{U}}]\subset(0,1)$ with $0<\tau_{\mathsf{L}}<\tau_{\mathsf{U}}<1$ and any given $m$, since $\tau_{\mathsf{L}}$ and $\tau_{\mathsf{U}}$ are not required to be symmetric, there are three cases for the relation between $[\tau_{\mathsf{L}},\tau_{\mathsf{U}}]$ and $[\frac{1}{m},1-\frac{1}{m}]$, that is, $[\frac{1}{m},1-\frac{1}{m}]\subset [\tau_{\mathsf{L}},\tau_{\mathsf{U}}]$, $[\tau_{\mathsf{L}},\tau_{\mathsf{U}}]\subseteq[\frac{1}{m},1-\frac{1}{m}]$, and $[\tau_{\mathsf{L}},\tau_{\mathsf{U}}]$ intersects with $[\frac{1}{m},1-\frac{1}{m}]$. We prove Proposition \ref{pro:asym2} for each of these three cases, respectively.

\textbf{Case 1:} $[\frac{1}{m},1-\frac{1}{m}]\subset[\tau_{\mathsf{L}},\tau_{\mathsf{U}}]$.
For $\tau\in[\tau_{\mathsf{L}},\tau_1)$,  we have $\hat{\bbeta}(\tau) = \tilde{\bbeta}_1$. Write
\[
\hat{\bbeta}(\tau)-\bbeta(\tau) = \tilde{\bbeta}_1-\bbeta(\tau) =\tilde{\bbeta}_1-\bbeta(\tau_1)+\bbeta(\tau_1)-\bbeta(\tau).
\]

For $\tau\in[\tau_j,\tau_{j+1})$, denote $\frac{\tau_{j+1}-\tau}{\tau_{j+1}-\tau_j}$ as $\alpha$, then we have $\hat{\bbeta}(\tau) = \alpha \tilde{\bbeta}(\tau_j)+ (1-\alpha)\tilde{\bbeta}_{j+1}$. Write
\begin{eqnarray}
	\lefteqn{\hat{\bbeta}(\tau)-\bbeta(\tau)}\\ \label{eq:beta_error}
	 &=&  \alpha \left(\tilde{\bbeta}_j-\bbeta(\tau_j)\right) + (1-\alpha)\left(\tilde{\bbeta}_{j+1}-\bbeta(\tau_{j+1})\right)
	+\alpha\left(\bbeta(\tau_j)-\bbeta(\tau)\right) + (1-\alpha)\left(\bbeta(\tau_{j+1})-\bbeta(\tau)\right). \nonumber
\end{eqnarray}

For $\tau\in[\tau_{m-1},\tau_{\mathsf{U}}]$, we have $\hat{\bbeta}(\tau) = \tilde{\bbeta}_{m-1}$. Write
\[
\hat{\bbeta}(\tau)-\bbeta(\tau) =  \tilde{\bbeta}_{m-1}-\bbeta(\tau) 
= \tilde{\bbeta}_{m-1}-\bbeta(\tau_{m-1})+\bbeta(\tau_{m-1})-\bbeta(\tau).
\]

Thus, recall that $F^{-1}_Y(\tau|\bx^*)=\bbeta(\tau)^\intercal\bx^*$, we then have 
\begin{eqnarray*}
	\lefteqn{ \sup_{\tau\in[\tau_{\mathsf{L}},\tau_{\mathsf{U}}]}\left|\hat{Q}(\tau|\bx^*)-F^{-1}_Y(\tau|\bx^*)\right|} \\ 
	&\le & \max_{\substack{\tau_{j}\in[\tau_{\mathsf{L}},\tau_{\mathsf{U}}] \\ 1\le j\le {m-1} }}\|\tilde{\bbeta}_j-\bbeta(\tau_j)\|\cdot\|\bx^*\|+ \sup\Bigg\{\sup_{\tau\in[\tau_{\mathsf{L}},\tau_{1})}\left|F_Y^{-1}(\tau_1|\bx^*)-F_Y^{-1}(\tau|\bx^*)\right|, \Bigg.\\
	&&\Bigg.\sup_{\substack{\tau\in[\tau_{j-1},\tau_{j+1}) \\ 2\le j\le {m-2}}} \left|F_Y^{-1}(\tau|\bx^*)-F_Y^{-1}(\tau_j|\bx^*)\right|,\sup_{\tau\in[\tau_{m-1},\tau_{\mathsf{U}}]}\left|F_Y^{-1}(\tau|\bx^*)-F_Y^{-1}(\tau_{m-1}|\bx^*)\right|\Bigg\} \\
	&\le & \max_{\substack{\tau_{j}\in[\tau_{\mathsf{L}},\tau_{\mathsf{U}}] \\ 1\le j\le {m-1}}}\|\tilde{\bbeta}_j-\bbeta(\tau_j)\|\cdot\|\bx^*\| +  \sup\Bigg\{\left|F_Y^{-1}(\tau_1|\bx^*)-F_Y^{-1}(\tau_{\mathsf{L}}|\bx^*)\right|,\Bigg.\\
	& &\Bigg.\sup_{1\le j\le {m-2}} \left|F_Y^{-1}(\tau_j|\bx^*)-F_Y^{-1}(\tau_{j+1}|\bx^*)\right|,\left|F_Y^{-1}(\tau_{\mathsf{U}}|\bx^*)-F_Y^{-1}(\tau_{m-1}|\bx^*)\right|\Bigg\}\\
	&\le & \max_{\substack{\tau_{j}\in[\tau_{\mathsf{L}},\tau_{\mathsf{U}}] \\ 1\le j\le {m-1}}}\|\tilde{\bbeta}_j-\bbeta(\tau_j)\|\cdot\|\bx^*\| + \frac{1}{m}\left(\inf_{\tau\in[\tau_{\mathsf{L}},\tau_{\mathsf{U}}]}f_Y(F_Y^{-1}(\tau|\bx^*)|\bx^*)\right)^{-1},
\end{eqnarray*}
where the last inequality is due to Assumption \ref{ass:cons2b}, the mean value theorem, and for any $\tau\in[\tau_{\mathsf{L}},\tau_{\mathsf{U}}]$,
\[
	\left(F_Y^{-1}(\tau|\bx^*)\right)'= \left(f_Y\left(F_Y^{-1}(\tau|\bx^*)\right)\right)^{-1}\le \left(\inf_{\tau\in[\tau_{\mathsf{L}},\tau_{\mathsf{U}}]}f_Y\left(F_Y^{-1}(\tau|\bx^*)\right)\right)^{-1}.
\]

\textbf{Case 2:} $[\tau_{\mathsf{L}},\tau_{\mathsf{U}}]\subseteq[\frac{1}{m},1-\frac{1}{m}]$.
 Compared with Case 1, this case is more tricky due to the problem caused by the two-side small intervals near the two endpoints $\tau_{\mathsf{L}}$ and $\tau_{\mathsf{U}}$, (e.g. $[\tau_{j},\tau_{\mathsf{U}}]$, if $\tau_{\mathsf{U}}\in[\tau_{j},\tau_{j+1}]$ for some $j$), where $\hat{\bbeta}(\tau)$ are interpolated by $\tilde{\bbeta}_j$ and $\tilde{\bbeta}_{j+1}$, however, $\tilde{\bbeta}_{j+1}$ is outside $[\tau_{\mathsf{L}},\tau_{\mathsf{U}}]$, we can't ensure its uniform convergence by Proposition \ref{pro:asym1}. To solve this problem, we compress each end of the interval $[\tau_{\mathsf{L}},\tau_{\mathsf{U}}]$ inward by $\frac{1}{m}$, that is, we only consider the interval $[\tau_{\mathsf{L}}+\frac{1}{m},\tau_{\mathsf{U}}-\frac{1}{m}]$, and it will converge to $(\tau_{\mathsf{L}},\tau_{\mathsf{U}})$ as $m\to \infty$. Then, notice that for any $\tau\in[\tau_{\mathsf{L}}+\frac{1}{m},\tau_{\mathsf{U}}-\frac{1}{m}]$, we can now always find $[\tau_j,\tau_{j+1})$ subject to $\tau\in[\tau_j,\tau_{j+1})\subseteq[\tau_{\mathsf{L}},\tau_{\mathsf{U}}]$. By equation~\eqref{eq:beta_error}, we have
\begin{eqnarray*}
\lefteqn{ \sup_{\tau\in[\tau_{\mathsf{L}}+\frac{1}{m},\tau_{\mathsf{U}}-\frac{1}{m}]}\left|\hat{Q}(\tau|\bx^*)-F^{-1}_Y(\tau|\bx^*)\right|} \\ 
	& \le & \max_{\substack{\tau_{j}\in[\tau_{\mathsf{L}},\tau_{\mathsf{U}}] \\ 1\le j\le {m-1}}}\|\tilde{\bbeta}_j-\bbeta(\tau_j)\|\cdot\|\bx^*\| + \sup\Bigg\{\sup_{\substack{\tau\in[\tau_{j},\tau_{j+1})\subseteq[\tau_{\mathsf{L}},\tau_{\mathsf{U}}] \\ 1\le j\le {m-1}}} \left|F_Y^{-1}(\tau_j|\bx^*)-F_Y^{-1}(\tau|\bx^*)\right|, \Bigg. \\
	& & \Bigg. \sup_{\substack{\tau\in[\tau_{j},\tau_{j+1})\subseteq[\tau_{\mathsf{L}},\tau_{\mathsf{U}}] \\ 1\le j\le {m-1}}} \left|F_Y^{-1}(\tau_{j+1}|\bx^*)-F_Y^{-1}(\tau|\bx^*)\right|\Bigg\}\\
	&\le & \max_{\substack{\tau_{j}\in[\tau_{\mathsf{L}},\tau_{\mathsf{U}}] \\ 1\le j\le {m-1}}}\|\tilde{\bbeta}_j-\bbeta(\tau_j)\|\cdot\|\bx^*\|+\sup_{\substack{\tau_{j}\in[\tau_{\mathsf{L}},\tau_{\mathsf{U}}] \\ 1\le j\le {m-1}}} \left|F_Y^{-1}(\tau_{j+1}|\bx^*)-F_Y^{-1}(\tau_{j}|\bx^*)\right| 
	\\
	&\le & \max_{\substack{\tau_{j}\in[\tau_{\mathsf{L}},\tau_{\mathsf{U}}] \\ 1 \le j\le {m-1}}}\|\tilde{\bbeta}_j-\bbeta(\tau_j)\|\cdot\|\bx^*\| + \frac{1}{m}\left(\inf_{\tau\in[\tau_{\mathsf{L}},\tau_{\mathsf{U}}]}f_Y(F_Y^{-1}(\tau|\bx^*)|\bx^*)\right)^{-1}.
\end{eqnarray*}
\textbf{Case 3:} $[\tau_{\mathsf{L}},\tau_{\mathsf{U}}]$ intersects with $[\frac{1}{m},1-\frac{1}{m}]$. This case is just a combination of Case 1 and Case 2. Following a similar proof, we can still have 
\begin{eqnarray}\label{pro2:eq01}
	\lefteqn{ \sup_{\tau\in[\tau_{\mathsf{L}}+\frac{1}{m},\tau_{\mathsf{U}}-\frac{1}{m}]}\left|\hat{Q}(\tau|\bx^*)-F^{-1}_Y(\tau|\bx^*)\right|} \nonumber\\ 
	&\le & \max_{\substack{\tau_{j}\in[\tau_{\mathsf{L}},\tau_{\mathsf{U}}] \\ 1\le j\le {m-1}}}\|\tilde{\bbeta}_j-\bbeta(\tau_j)\|\cdot\|\bx^*\| + \frac{1}{m}\left(\inf_{\tau\in[\tau_{\mathsf{L}},\tau_{\mathsf{U}}]}f_Y(F_Y^{-1}(\tau|\bx^*)|\bx^*)\right)^{-1}.
\end{eqnarray}

Thus, whichever case it is, we all have \eqref{pro2:eq01}.
Hence, as $n, m \rightarrow \infty$, by Proposition \ref{pro:asym1}, we have
$$
\sup _{\tau \in\left(\tau_{\mathsf{L}}, \tau_{\mathsf{U}}\right)}\left|\hat{Y}\left(\tau| \bx^*\right)-F_Y^{-1}(\tau \mid \bx^*)\right| \stackrel{\mathbb{P}}{\rightarrow} 0.
\halmos
$$

\subsection{Proof of Theorem \ref{thm:weakconvergence}}
We prove the convergence in distribution of ${Y}(\bx^*)$ by Lemma \ref{lemma:van}. Suppose $h$ is any given bounded, Lipschitz continuous function on $\mathcal Y$ such that there exist positive constants $M_h$ and $L_h$ that, $|h(y)|\le M_h$, for all $y\in\mathcal Y,$
and $|h(y)-h(y')|\le L_h|y-y'|$, for all $y$ and $y'$.  
Then, it is sufficient to show that 
\[
\mathbb{E}h(\hat{Q}(U|\bx^*))\to \mathbb{E}h(F^{-1}_Y(U|\bx^*)),
\]
as $n, m\to\infty$, where the distribution of $F^{-1}_Y(U|\bx^*)$ given $\bx^*$ is the conditional distribution $F_Y(\cdot|\bx^*)$.

For any given $\epsilon_1>0 $, by Lemma \ref{lemma:I3}, given $\tau_{\mathsf{L}}=\frac{\epsilon_1}{10(M_h\lor 1)}$, $\tau_{\mathsf{U}}=1-\frac{\epsilon_1}{10(M_h\lor 1)}$, for any $\epsilon_2>0 $, there exist sufficiently large $N(\epsilon_1,\epsilon_2),M(\epsilon_1,\epsilon_2)$, such that for any $n\ge N(\epsilon_1,\epsilon_2)$, $m\ge M(\epsilon_1,\epsilon_2)$, we have
\begin{equation}\label{thm01:eq01}
\mathbb{E} \left[\left|\left(h(\hat{Q}(U|\bx^*))-h(F^{-1}_Y(U|\bx^*))\right)\right|\mathbb{I}\{U\in I_3\}\right]\le \frac{\epsilon_2}{5}.
\end{equation}
Then, take sufficiently large $M^{\prime}(\epsilon_1,\epsilon_2)=M(\epsilon_1,\epsilon_2)\lor\frac{10((M_h\lor 1))}{\epsilon_1}$ such that for any $n\ge N(\epsilon_1,\epsilon_2)$, $m\ge M^{\prime}(\epsilon_1,\epsilon_2)$, we have that $\frac{1}{m}\le \frac{\epsilon_1}{10(M_h\lor 1)}$ and \eqref{thm01:eq01}. Notice that by the analysis of Proposition \ref{pro:asym2}, \eqref{thm01:eq01} keeps valid no matter what relationship between $[\tau_{\mathsf{L}},\tau_{\mathsf{U}}]$ and $[\frac{1}{m},1-\frac{1}{m}]$ is. 

Since \eqref{thm01:eq01} is satisfied for any $\epsilon_1,\epsilon_2>0 $, let $\epsilon_2=\epsilon_1$,
then we have the following statement: 
for any given bounded Lipschitz continuous function $h$, any  
$\epsilon_1>0 $, $\tau_{\mathsf{L}}=\frac{\epsilon_1}{10(M_h\lor 1)}$, and $\tau_{\mathsf{U}}=1-\frac{\epsilon_1}{10(M_h\lor 1)}$, there exist sufficiently large $N= N(\epsilon_1,\epsilon_1),M^{\prime}= M'(\epsilon_1,\epsilon_1)$, such that for any $n\ge N$, $m\ge M^{\prime}$, we have that $\frac{1}{m}\le \frac{\epsilon_1}{10(M_h\lor 1)}$ and \eqref{thm01:eq01}.
Hence, it is straightforward to see
\begin{eqnarray*}
	\lefteqn{
	\left|\mathbb{E}h(\hat{Q}(U|\bx^*))-\mathbb{E}h(F^{-1}_Y(U|\bx^*))\right| }\\
	& \le &  \mathbb{E} \left[\left|\left(h(\hat{Q}(U|\bx^*))-h(F^{-1}_Y(U|\bx^*))\right)\right|\mathbb{I}\{U\in I_1\}\right]+\mathbb{E} \left[\left|\left(h(\hat{Q}(U|\bx^*))-h(F^{-1}_Y(U|\bx^*))\right)\right|\mathbb{I}\{U\in I_2\}\right] \\
	&+&
	 \mathbb{E} \left[\left|\left(h(\hat{Q}(U|\bx^*))-h(F^{-1}_Y(U|\bx^*))\right)\right|\mathbb{I}\{U\in I_3\}\right] 
	 + \mathbb{E} \left[\left|\left(h(\hat{Q}(U|\bx^*))-h(F^{-1}_Y(U|\bx^*))\right)\right|\mathbb{I}\{U\in I_4\}\right]\\
	&+& \mathbb{E} \left[\left|\left(h(\hat{Q}(U|\bx^*))-h(F^{-1}_Y(U|\bx^*))\right)\right|\mathbb{I}\{U\in I_5\}\right]\\
	&\le & \frac{8M_h\cdot\epsilon_1}{10(M_h\lor 1)} +\frac{\epsilon_1}{5}
	\le \epsilon_1,
\end{eqnarray*}
which completes the proof of Theorem \ref{thm:weakconvergence_a}. Theorem \ref{thm:weakconvergence_b} can be readily derived from Theorem \ref{thm:weakconvergence_a} and Assumption \ref{ass:cons2a}, as the convergence in distribution and the continuity of the limit imply uniform convergence of the CDF \citep[Lemma~2.11]{van2000}. \halmos

\subsection{Proof of Proposition \ref{pro:asym3}}
By Lemma \ref{lemma:brown}, there exists a $p$-dimensional standard Brownian bridge $\bf B$ on $[0,1]$ such that 
\[
\sqrt{n}D_0^{-1/2}D_1(\tau)\left[\hat\bbeta(\tau)-\bbeta(\tau)\right]\Rightarrow{\bf B}(\tau)
\]
uniformly for $\tau\in[\tau_\ell,\tau_u]$, this implies that
\[
\sqrt{n}D_0^{-1/2}D_1(\tau_j)\left[\hat\bbeta_j-\bbeta(\tau_j)\right]=O_{\mathbb P}(1)
\]
uniformly for all $\tau_j\in[\tau_\ell,\tau_u],j=1,\ldots,m$.
Thus, by Assumptions (2.b) and (3.a), we have $\sup_{\tau\in[\tau_\ell,\tau_u]}\left\|D^{-1}_1(\tau)\right\|<\infty$, which implies
\[
\max_{\substack{\tau_{j}\in[\tau_\ell,\tau_u] \\ 1\le j\le m-1}}\sqrt{n}\left\|\hat\bbeta_j-\bbeta(\tau_j)\right\|=O_{\mathbb P}(1).
\halmos
\]

\section{Numerical Demonstration of QRGMM-R}\label{e-sec:QRGMM-R-numerical}

We demonstrate QRGMM-R's performance using Test Problem 2 from Section \ref{sec:synthetic}, comparing it with QRGMM. 
Test Problem 2 exhibits more frequent quantile crossing than Test Problem 1 due to model misspecification, as its true conditional quantile function violates the linear quantile regression assumptions.
Over 100 replications, the quantile crossing frequency is $0.0695 ~(\pm 0.0202)$.

We evaluate QRGMM-R using the experimental design from Section \ref{subsec:syn_compare}, examining sample means, standard deviations, Wasserstein distances, and KS statistics. 
Results appear in Table \ref{table:qqrgmm-r} and Figure \ref{fig_QC_WDKS}, 
showing minimal distributional differences between QRGMM and QRGMM-R.

\begin{table}[t]
	\TABLE{Generated Samples Approximating Distribution of $Y$. \label{table:qqrgmm-r}}
	{\begin{tabular}{ccccccccccccccccc}
			\toprule 
			\multirow{2}{*}{} && \multicolumn{3}{c}{Sample Mean}    && \multicolumn{3}{c}{Sample SD} \\ 
			\cmidrule(l){3-5} \cmidrule(l){7-9}
			&& {Mean}       && {SD}          && {Mean}          && {SD} 
			\\ 
			\cmidrule(l){1-1} \cmidrule(l){3-3} \cmidrule(l){5-5} \cmidrule(l){7-7} \cmidrule(l){9-9} 
			Truth       && 0.796524    && 0.048984  && 13.991365    && 0.050623 \\ 
			QRGMM             && 0.770990    && 0.298282 && 10.316667    && 0.321973\\
			QRGMM-R    && 0.770992    && 0.298280  && 10.316670    && 0.321972  \\
			\bottomrule
		\end{tabular}
	}
	{}
\end{table}

\begin{figure}[ht]	\FIGURE{\includegraphics[width=0.75\textwidth]{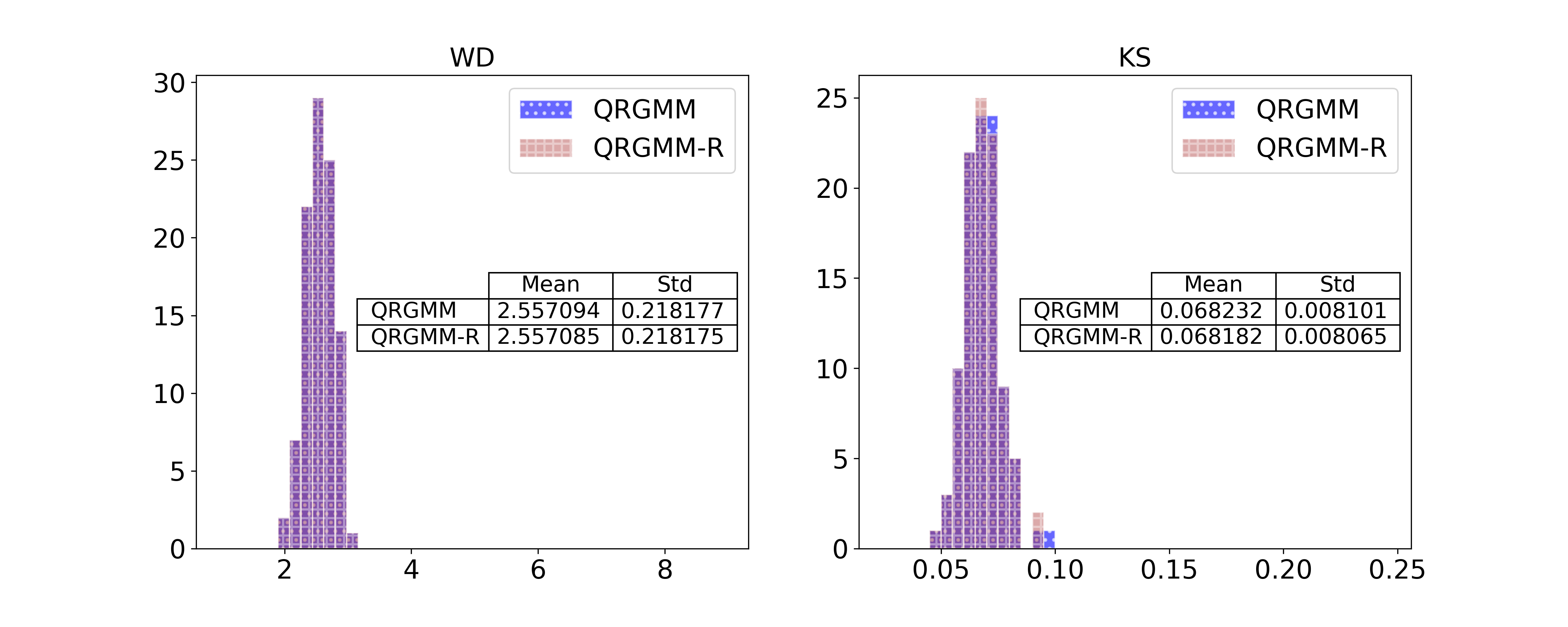}}
	{Generated Samples Approximating Distribution of $Y$: WD and KS.  \label{fig_QC_WDKS} }
	{}
\end{figure}

In terms of computational speed,
QRGMM-R generates $10^5$ samples in 0.0040 seconds compared to 0.0039 seconds for QRGMM, averaged over 100 replications. 
The rearrangement step thus adds minimal computational overhead while preserving the distributional properties of QRGMM.

	\section{Asymptotic Analysis of QRGMM-R} \label{e-sec:convergence}
	
	In this section, we first present \cite{neocleous2008monotonicity}'s findings on quantile crossing in linear quantile regression models. 
	We then establish the convergence in distribution of QRGMM-R. Our analysis builds on results from \cite{neocleous2008monotonicity} and \cite{chernozhukov2010quantile}.

	\subsection{Frequency of Quantile Crossing} \label{e-sec:no-crossing}
	We first list the assumptions required in \cite{neocleous2008monotonicity}, 
	which are stronger than Assumptions \ref{ass:cons2b} and \ref{ass:cons3b}. 
	
	\begin{assumption}\label{ass:neo}	
		\begin{enumerate}[label=(EC.3.\alph*), left=0pt]
			\item For any given interval $[\tau_{\mathsf{L}},\tau_{\mathsf{U}}]\subset(0,1)$ with $0<\tau_{\mathsf{L}}<\tau_{\mathsf{U}}<1$, and any bounded $\mathcal{X}$, there exist constants $a, b, c$ with $a>0, b<\infty$, and $c<\infty$ such that
			$$
			a \leq f_Y(F_Y^{-1}(\tau|\bx)|\bx) \leq b, \quad\left|f_Y^{\prime}(F_Y^{-1}(\tau|\bx)|\bx)\right| \leq c
			$$
			uniformly for $\bx \in \mathcal{X}, \tau \in [\tau_{\mathsf{L}},\tau_{\mathsf{U}}]$. \label{ass:neoa}	
			\item $\left\|\bx_i\right\| \leq d$ for some constant $d$ uniformly in $i=1, \ldots, n$.\label{ass:neob}	
		\end{enumerate} 
	\end{assumption}

	\begin{lemma}[\cite{neocleous2008monotonicity}]\label{lemma:neo} 
		Under Assumptions \ref{ass:cons1}, \ref{ass:cons2a}, \ref{ass:cons3a}, and \ref{ass:neo}, let $m$ satisfy $\limsup\limits_{n\to\infty} n^{\eta} / m>0$ and $\liminf\limits_{n\to\infty} n^{1 / 2} / (m\log n)>0 $ for some $\eta>0$. Then with probability tending to one, $\hat{Q}(\tau|\bx)$ is strictly monotonic uniformly on $\{\bx\mid\|\bx\| \leq d\}$ for $\tau \in [\tau_{\mathsf{L}}+\frac{1}{m},\tau_{\mathsf{U}}-\frac{1}{m}]$, and
		$$
		\sqrt{n}\left(\hat{\bbeta}(\tau)-\tilde{\bbeta}(\tau)\right)=O\left(\sqrt{n}/(m^2) \right)+O_{\mathbb P}\left(\left(\log n/m\right)^{1 / 2}\right)+O_{\mathbb P}\left(n^{-1 / 4}(\log n)^{1 / 2}\right)
		$$
		uniformly for $\tau \in [\tau_{\mathsf{L}}+\frac{1}{m},\tau_{\mathsf{U}}-\frac{1}{m}]$.
	\end{lemma}
	
	The implications of Lemma \ref{lemma:neo} are twofold. First, strict monotonicity of $\hat{Q}(\tau|\bx)$ over $(\tau_{\mathsf{L}},\tau_{\mathsf{U}})$ is guaranteed when $m$ grows slower than $\sqrt{n}$. Second, when $m$ grows faster than $n^{1/4}$, $\hat{\bbeta}(\tau)$ maintains the same first-order asymptotic Bahadur representation as $\tilde{\bbeta}(\tau)$.

	\subsection{Convergence in Distribution of QRGMM-R} \label{e-sec:convergence_in_distribution}

	\begin{assumption}\label{ass:cher}	
		$Q: (0,1)\times \mathcal{X} \mapsto \mathbb{R}$ is a continuously differentiable function in both arguments, and $\mathcal{X}$ is a compact subset of $\mathbb{R}^d$.
	\end{assumption}
	
	\begin{assumption}\label{ass:monotonicity}	
		$\tau \mapsto Q(\tau | \bx)$ has $\partial_\tau Q(\tau | \bx)>0$ for each $(\tau, \bx) \in (0,1)\times \mathcal{X}$.
	\end{assumption}
	
	Under Assumption \ref{ass:cons1}, the monotonicity of $Q(\tau|\bx)$ in $\tau$ (Assumption \ref{ass:monotonicity}) follows immediately.

	\begin{lemma}[\cite{chernozhukov2010quantile}]\label{lemma:cher} 
		Let $\mathcal{U}$ be $(0,1)$ or its subinterval $(a,b)$, and let $\ell^{\infty}(\mathcal{U}\times\mathcal{X})$ denote the space of bounded measurable functions $h:\mathcal{U}\times\mathcal{X} \mapsto \mathbb{R}$. Consider an estimated quantile curve $\hat{Q}$ in $\ell^{\infty}(\mathcal{U}\times\mathcal{X})$. 
		Under Assumptions \ref{ass:cher} and \ref{ass:monotonicity}, if
		\[a_n\left(\hat{Q}(\tau | \bx)-Q(\tau | \bx)\right) \Rightarrow G(\tau | \bx)
		\]
		in $\ell^{\infty}(\mathcal{U}\times\mathcal{X})$ as a stochastic process indexed by $(\tau, \bx) \in \mathcal{U} \times \mathcal{X}$, where 
		$a_n$ is a sequence of constants going to infinity as $n \rightarrow \infty$, and 
		$G(\tau | \bx)$ is a stochastic process with continuous paths, then
		\[a_n\left(\hat{Q}^{\mathsf{R}}(\tau | \bx)-Q(\tau | \bx)\right) \Rightarrow G(\tau | \bx)
		\]
		uniformly over $\mathcal{U}\times\mathcal{X}$, establishing that the rearranged quantile curve maintains the same first-order asymptotic distribution as the original estimate.
	\end{lemma}

	Note that the output random variable of QRGMM-R is $\hat{Y}^{\mathsf{R}}(\bx^*) :=\hat{Q}^{\mathsf{R}}(U | \bx^*)$, where $U \sim \mathsf{Unif}(0,1)$. 
	By Lemma \ref{lemma:brown}, under Assumptions \ref{ass:cons1}-\ref{ass:cons3}, we know that for any given interval $[\tau_{\mathsf{L}},\tau_{\mathsf{U}}]\subset(0,1)$, we have that  $\sqrt{n}\left[\tilde{\bbeta}(\tau)-\bbeta(\tau)\right]$ converge in distribution to a p-dimensional Brownian bridge uniformly over $\tau \in [\tau_{\mathsf{L}},\tau_{\mathsf{U}}]$. Then under Assumption \ref{ass:neo}, according to Lemma \ref{lemma:neo}, as long as $m$ increases faster than $n^{1/4}$, $\sqrt{n}\left[\hat{Q}(\tau|\bx^*)-Q(\tau|\bx^*)\right]$ also converge to a Gaussian process uniformly over $[\tau_{\mathsf{L}}+\frac{1}{m},\tau_{\mathsf{U}}-\frac{1}{m}] \times \mathcal{X} $, denoted as $G(\tau|\bx^*)$. Then by Lemma \ref{lemma:cher}, we have the following corollary.

	\begin{corollary}\label{corollary2} 
		Under Assumptions \ref{ass:cons1}, \ref{ass:cons2a}, \ref{ass:cons3a}, \ref{ass:neo}, and \ref{ass:cher}, we have 
		$$
		\sqrt{n}\left[\hat{Q}^{\mathsf{R}}(\tau|\bx^*)-Q(\tau|\bx^*)\right] \Rightarrow G(\tau|\bx^*)
		$$ uniformly over $[\tau_{\mathsf{L}}+\frac{1}{m},\tau_{\mathsf{U}}-\frac{1}{m}] \times \mathcal{X} $. 
		
	\end{corollary}

	This corollary implies that all three estimators $\tilde{Q}(\tau|\bx^*)$, $\hat{Q}(\tau|\bx^*)$, and $\hat{Q}^{\mathsf{R}}(\tau|\bx^*)$ share the same first-order asymptotic distribtion. This further indicates that the asymptotic analysis for $\hat{Y}(\bx^*)$ remains unaffected by the rearrangement operation of the new algorithm. Then the convergence in distribution of $\hat{Y}^{\mathsf{R}}(\bx^*)$, which is restated in a more rigorous form in the following Theorem \ref{thm3_rig}, can be easily obtained through Corollary \ref{corollary2}. As Corollary \ref{corollary2} implies that for any given $\bx^*\in\mathcal{X}$, as $n, m \rightarrow \infty$, 
	\[
	\sup_{\tau\in I_3}\left|\hat{Q}^{\mathsf{R}}(\tau|\bx^*)-F^{-1}_Y(\tau|\bx^*)\right| \stackrel{\mathbb{P}}{\rightarrow} 0,
	\]
	which is aligned with Proposition \ref{pro:asym2}, then according to Lemma \ref{lemma:I3}, we can prove Theorem \ref{thm3_rig} (a rigorous restatement of Theorem~\ref{Thm03}) following the same way of Theorem \ref{thm:weakconvergence}.
	
	\begin{theorem}\label{thm3_rig} 
		Under Assumptions \ref{ass:cons1}, \ref{ass:cons2a}, \ref{ass:cons3a}, \ref{ass:neo}, and \ref{ass:cher}, the following results hold:
		\begin{enumerate}[label=(2.\alph*), left=0pt]
			\item For any $\bx^*\in \mathcal{X}$, $\hat{Y}^{\mathsf{R}}(\bx^*) \Rightarrow Y(\bx^*)$ as $n,m\to\infty$ . 	
			\item For any $\bx^*\in \mathcal{X}$,
			\[
			\sup_{y\in\left(-\infty,\infty\right)}\left|F_{\hat{Y}^{\mathsf{R}}}(y|\bx^*)-F_Y(y|\bx^*)\right| \to 0,
			\]
			as $n,m\to\infty$, 
			where $F_{\hat{Y}^{\mathsf{R}}}(y|\bx^*)$ is the conditional CDF of $\hat{Y}^{\mathsf{R}}(\bx^*)$ given $\bx^*$. 
		\end{enumerate}
	\end{theorem}

\section{Supplementary Materials for Section~\ref{sec:extension} }
\subsection{Multi-output QRGMM Algorithm}

To extend the QRGMM framework to multidimensional outputs, we leverage the conditional distribution method, which sequentially generates univariate conditional distributions while preserving the use of single-output quantile regression models, as detailed in Algorithm \ref{alg:Mul-QRGMM}. This approach avoids the complexities of joint quantile definitions and maintains the framework’s core structure.

\begin{algorithm}[t]
	\caption{Multi-output QRGMM}\label{alg:Mul-QRGMM}
	\begin{algorithmic}
		\State \textbf{Offline Stage:}
		\begin{itemize}
			\item Collect a dataset $\{(\bx_i,\by_i):i=1,\ldots,n\}$, and choose a quantile regression model $Q_l(\tau|\bx, \by_{[l-1]})$ for the conditional quantile function $F_{Y_l}^{-1}(\tau| \bx, \by_{[l-1]})$ for each $l=1,\ldots,d$.
			\item Choose a positive integer $m$ and create the grid $\{\tau_j \coloneqq j/m: j=1,\ldots,m-1\}$.
			\item For each $j=1,\ldots,m-1$ and $l=1,\ldots,d$, fit the model $Q_l(\tau|\bx, \by_{[l-1]})$ with quantile level $\tau_j$ to the dataset, yielding  $\tilde{Q}_l(\tau_j|\bx, \by_{[l-1]})$. \end{itemize}			
		
		\State \textbf{Online Stage:} 
		\begin{itemize}
			\item Observe the covariates $\bx=\bx^*$. 
			\item Generate $\{u_{l,k}:l=1,\ldots,d, k=1,\ldots,K\}$ independently from $\mathsf{Unif}(0,1)$.
			\item For each $l=1,\ldots,d$ and $k=1,\ldots,K$, calculate 
			$\hat{Y}_{l,k} = \hat{Q}_l(u_{l,k}|\bx^*, \hat{Y}_{1,k}, \ldots, \hat{Y}_{l-1,k})$, where $\tilde{Q}_l(\tau|\bx, \by_{[l-1]})$ is the linear interpolator of $\{(\tau_j, \tilde{Q}_l(\tau_j|\bx, \by_{[l-1]})):j=1,\ldots,m-1\}$.
			\item Output: $\{\hat{\bY}_{k}:k=1,\ldots,K)\}$, where $\hat{\bY}_{k} = (\hat{Y}_{1,k}, \ldots, \hat{Y}_{d,k})$.
		\end{itemize}
		
	\end{algorithmic}
\end{algorithm}

\subsection{Multi-output Simulation for a Bank Queueing System} \label{sec:bank-sym}

We demonstrate the performance of $\text{QRGMM}^{+}$, which integrates \cite{cannon2011quantile}'s neural network quantile regression model into the multi-output QRGMM framework (Algorithm \ref{alg:Mul-QRGMM}), using a bank queueing simulation. 
Operating from 9:00 AM to 5:00 PM, this simulation focuses on transient rather than steady-state behavior, providing a complex multidimensional testing environment.

The simulator models two customer types with preemptive priority service: priority customers interrupt the service of normal customers and must be completely served before normal customers resume service. Customer arrivals follow Poisson processes with type-specific rates, and customers may renege (prematurely exit) if their waiting time exceeds a Gamma-distributed threshold. The bank operates six service counters, each with an individual queue and exponentially distributed service times at counter-specific rates. The simulator takes ten-dimensional input parameters $\bx\in\mathbb{R}^{10}$ comprising two customer arrival rates, six counter service rates, and two reneging distribution parameters. It generates five output metrics $\bY(\bx)=\left(Y_1(\bx),Y_2(\bx),Y_3(\bx),Y_4(\bx),Y_5(\bx)\right)$: average system time and renege rate for both customer types, plus maximum counter overtime after closing. The simulation is implemented using R's {\it simmer} discrete-event-simulation package.

We train $\text{QRGMM}^{+}$ and other models using $n=10^4$ samples $\{(\bx_i,\by_{i})\}_{i=1}^n$, where $\bx$ follows a uniform distribution over a hypercube. For evaluation, we generate $K=10^5$ samples at a fixed $\bx^*$ from each trained metamodel and the simulator, comparing their conditional marginal distributions of $Y_l$ given $\bx^*$ for each dimension $l=1,\ldots,5$ in Figure \ref{fig:conditional_marginal_distributions}.
The conditional marginal distributions of samples from $\text{QRGMM}^{+}$, Diffusion, and RectFlow notably match the true distributions more closely than those from CWGAN across all dimensions.

\begin{figure}[ht]
	\FIGURE{
		\includegraphics[width=\textwidth]{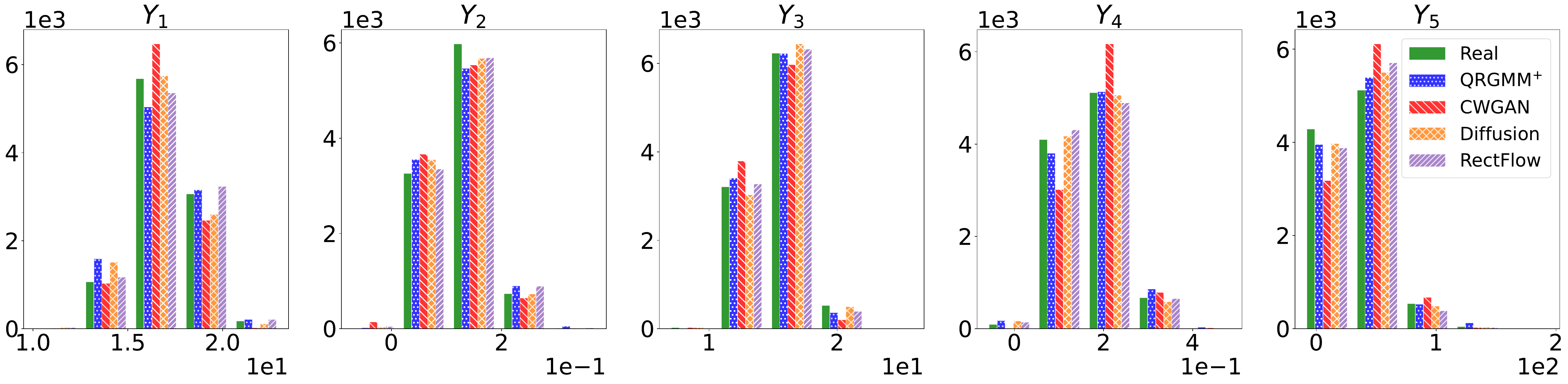} }
	{Generated Samples Approximating Distribution of $Y_l$ given $\bx^* =  \left( \frac{1}{4}, \frac{1}{8}, \frac{1}{15}, \frac{1}{16}, \frac{1}{17}, \frac{1}{19}, \frac{1}{11}, \frac{1}{13}, 18, 12 \right)$. \label{fig:conditional_marginal_distributions}}
	{}
\end{figure}

To evaluate unconditional distributions, we use each trained metamodel to produce a dataset $\{(\bx^{\prime}_k,\hat{\by}^{\prime}_k)\}_{k=1}^K$ and compare it against a test dataset $\{(\bx^{\prime}_k,\by^{\prime}_k)\}_{k=1}^K$ generated from the simulator. 
For each dimension $l=1,\ldots,5$, we calculate the sample mean and sample standard deviation of $\{\hat{y}^{\prime}_{l,k}\}_{k=1}^K$ or $\{y^{\prime}_{l,k}\}_{k=1}^K$.
We also compute the Wasserstein distance between the multi-dimensional samples from each metamodel and the simulator using the \texttt{POT} Python library for optimal transport computations. %

Across 100 replications, we calculate means and standard deviations of all summary statistics (sample mean, sample standard deviation, and Wasserstein distance).  As shown in Table~\ref{table_bank_MSD}, $\text{QRGMM}^{+}$ consistently provides the closest approximation to the true first two moments across all output dimensions, outperforming other models. Moreover, $\text{QRGMM}^{+}$ yields substantially smaller Wasserstein distances with lower variability (Figure~\ref{fig_bank_WD}), indicating a more accurate and stable recovery of the full conditional distribution for complex stochastic systems.

\begin{table}[ht]
	\TABLE{Generated Samples Approximating Distribution of $\bY$: Mean and Standard Deviation.
		\label{table_bank_MSD}}
	{
		\begin{tabular}{@{}lcccccccccc@{}}
			\toprule
			& \multicolumn{2}{c}{$Y_1$}
			& \multicolumn{2}{c}{$Y_2$}
			& \multicolumn{2}{c}{$Y_3$}
			& \multicolumn{2}{c}{$Y_4$}
			& \multicolumn{2}{c}{$Y_5$} \\
			\cmidrule(lr){2-3} \cmidrule(lr){4-5} \cmidrule(lr){6-7}
			\cmidrule(lr){8-9} \cmidrule(lr){10-11}
			& Mean & SD & Mean & SD & Mean & SD & Mean & SD & Mean & SD \\
			\midrule
			\multicolumn{11}{c}{\textbf{Sample Mean}} \\
			\midrule
			Truth
			& 18.2215 & 0.0193 & 0.3794 & 0.0016 & 15.2626 & 0.0189 & 0.3316 & 0.0011 & 41.4609 & 0.2125 \\
			$\text{QRGMM}^{+}$
			& 18.2202 & 0.0239 & 0.3794 & 0.0018 & 15.2605 & 0.0219 & 0.3316 & 0.0014 & 41.3725 & 0.2592 \\
			CWGAN
			& 18.2490 & 0.1762 & 0.3830 & 0.0186 & 15.2804 & 0.1806 & 0.3309 & 0.0134 & 41.5027 & 2.0023 \\
			Diffusion
			& 18.2290 & 0.0745 & 0.3794 & 0.0042 & 15.2237 & 0.1161 & 0.3311 & 0.0049 & 39.1964 & 1.1825 \\
			RectFlow
			& 18.2163 & 0.0436 & 0.3790 & 0.0031 & 15.2213 & 0.0796 & 0.3310 & 0.0034 & 39.0578 & 0.7876 \\
			\midrule
			\multicolumn{11}{c}{\textbf{Sample SD}} \\
			\midrule
			Truth
			& 2.1663 & 0.0163 & 0.1769 & 0.0008 & 1.9447 & 0.0130 & 0.1211 & 0.0008 & 19.4521 & 0.2214 \\
			$\text{QRGMM}^{+}$
			& 2.1614 & 0.0187 & 0.1767 & 0.0009 & 1.9242 & 0.0200 & 0.1208 & 0.0009 & 18.9545 & 0.2535 \\
			CWGAN
			& 2.1203 & 0.1498 & 0.1739 & 0.0106 & 1.9015 & 0.1147 & 0.1186 & 0.0090 & 18.7522 & 1.4463 \\
			Diffusion
			& 2.1553 & 0.0350 & 0.1767 & 0.0016 & 1.9143 & 0.0291 & 0.1210 & 0.0017 & 18.4099 & 0.7548 \\
			RectFlow
			& 2.1615 & 0.0315 & 0.1769 & 0.0016 & 1.9194 & 0.0336 & 0.1208 & 0.0017 & 17.9479 & 0.4712 \\
			\bottomrule
		\end{tabular}
	}
	{}
\end{table} 

\begin{figure}[ht]
	\FIGURE{\includegraphics[width=0.5\textwidth]{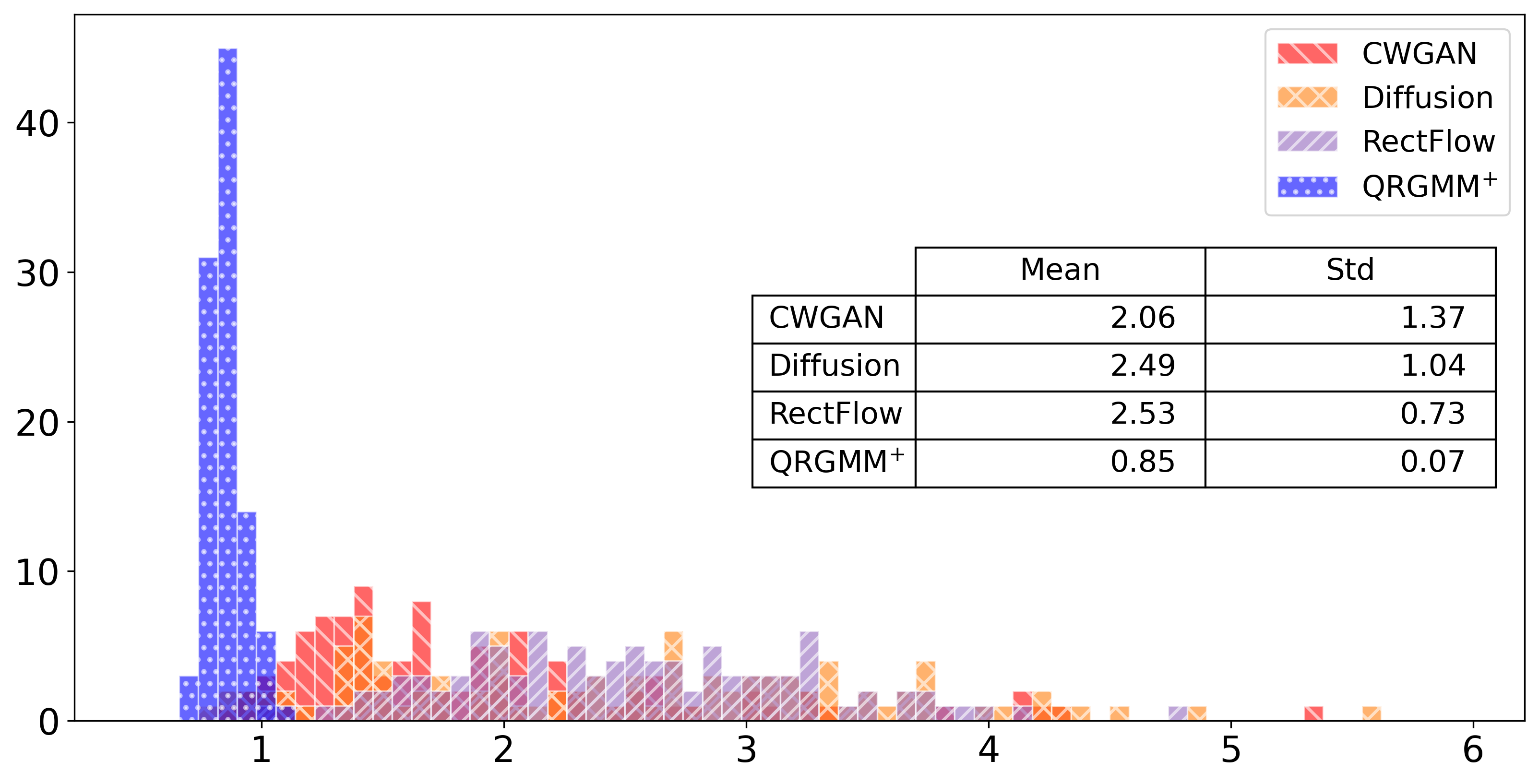}}
	{Generated Samples Approximating Distribution of $\bY$: Wassserstein Distance.  \label{fig_bank_WD} }
	{}
\end{figure}

Lastly, we evaluate online sample generation speed across 100 replications. $\text{QRGMM}^{+}$ requires on average $0.1426$ seconds (std.\ $0.0237$) to generate $10^4$ samples, which is slower than CWGAN ($0.0434 \pm 0.0027$ seconds) but comparable to Diffusion ($0.1801 \pm 0.0876$ seconds) and RectFlow ($0.0947 \pm 0.0010$ seconds). The slower speed of $\text{QRGMM}^{+}$ stems from its sequential generation process across output dimensions. %
Nevertheless, all generative metamodels considered here operate well within real-time constraints for this five-dimensional setting. Further improving the computational efficiency of $\text{QRGMM}^{+}$ for higher-dimensional output spaces remains an important direction for future research.

\section{Supplementary Materials for Section~\ref{sec:num} }

This subsection complements Section~\ref{treatment_effects} by providing additional results on the unconditional distributions of $Y_1$ and $Y_2$ obtained from generated samples. Table~\ref{table:EC_vs_MSD} and Figure~\ref{fig_ECS_vs_hist_WDKS_Y12} summarize the performance of all models in approximating these distributions. The results are consistent with the findings in Section~\ref{subsec:syn_compare}, demonstrating that QRGMM achieves better accuracy and stability across multiple summary statistics.

\begin{table}[t]
	\TABLE{Generated Samples Approximating Distribution of $Y_1$ and $Y_2$: Mean and Standard Deviation.  \label{table:EC_vs_MSD} }
	{\begin{tabular}{ccccccccccccccccc}
			\toprule 
			\multirow{3}{*}{} && \multicolumn{7}{c}{Sample Mean} && \multicolumn{7}{c}{Sample SD}\\
			\cmidrule(l){3-9} 
			\cmidrule(l){11-17} 
			&& \multicolumn{3}{c}{$Y_1$}     && \multicolumn{3}{c}{$Y_2$} && \multicolumn{3}{c}{$Y_1$}     && \multicolumn{3}{c}{$Y_2$}            \\ 
			\cmidrule(l){3-5} \cmidrule(l){7-9} \cmidrule(l){11-13} \cmidrule(l){15-17} 
			&& {Mean}       && {SD}          && {Mean}          && {SD} 
			&& {Mean}       && {SD}          && {Mean}          && {SD}         \\ 
			\cmidrule(l){1-1} \cmidrule(l){3-3} \cmidrule(l){5-5} \cmidrule(l){7-7} \cmidrule(l){9-9} \cmidrule(l){11-11} \cmidrule(l){13-13} \cmidrule(l){15-15} \cmidrule(l){17-17} 
			
			Truth              
			&& $15.6000$ && $0.0894$ && $15.6290$ && $0.0921$ 
			&& $9.5288$  && $0.0553$ && $9.5357$  && $0.0533$ \\ 
			
			QRGMM             
			&& $15.5018$ && $0.1227$ && $15.5427$ && $0.1178$ 
			&& $9.5026$  && $0.0821$ && $9.5127$  && $0.0747$  \\ 
			
			CWGAN             
			&& $15.4213$ && $2.7809$ && $15.8217$ && $2.6824$ 
			&& $8.0462$  && $1.7469$ && $8.0826$  && $1.7796$  \\ 
			
			Diffusion
			&& $15.7320$ && $0.2996$ && $15.7170$ && $0.2859$
			&& $9.9453$  && $0.1748$ && $9.9406$  && $0.1730$ \\
			
			RectFlow
			&& $15.7210$ && $0.2738$ && $15.7563$ && $0.2511$
			&& $9.9604$  && $0.1845$ && $9.9406$  && $0.1712$ \\
			
			\bottomrule
		\end{tabular}
	}
	{}
\end{table}

\begin{figure}[ht]
	\FIGURE{
		$\begin{array}{c}     
			\includegraphics[width=0.495\linewidth,height=0.22\linewidth]{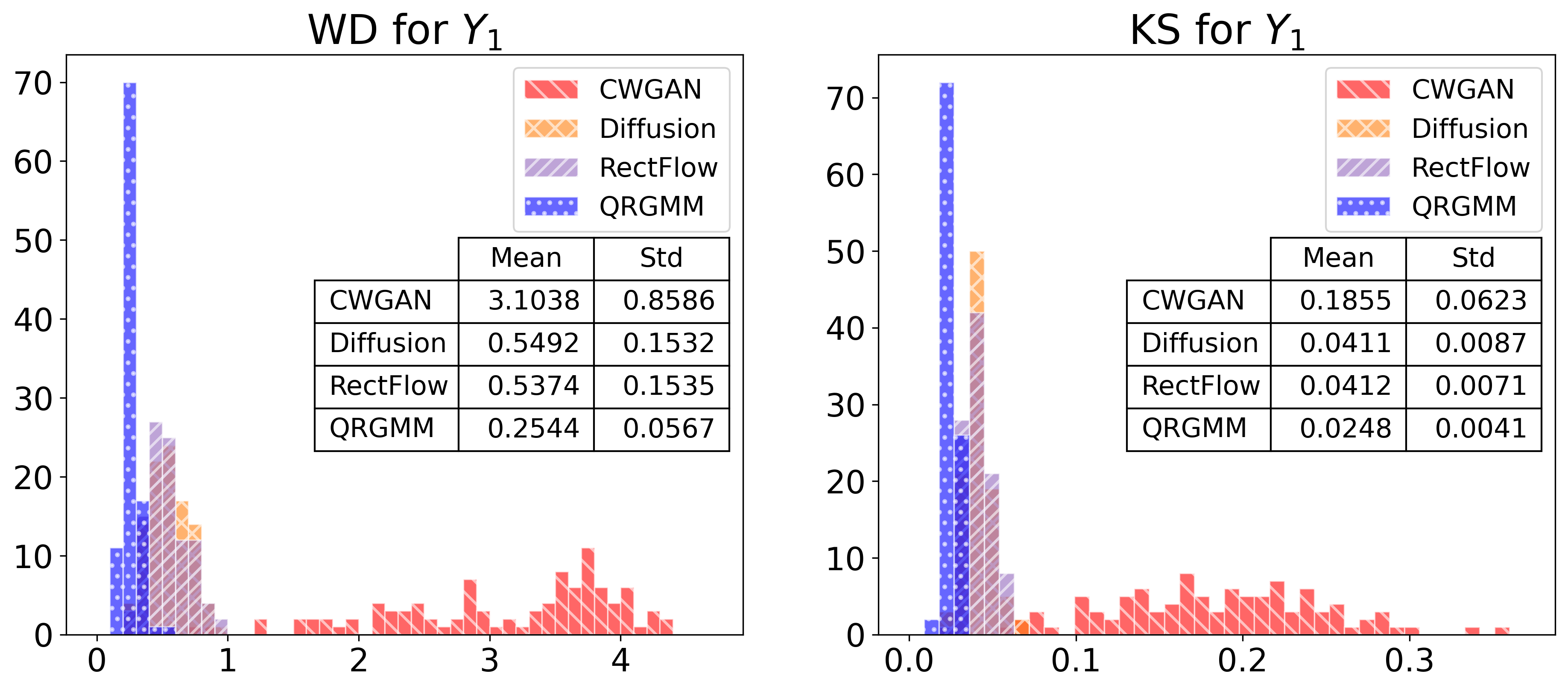} 
			\includegraphics[width=0.495\linewidth,height=0.22\linewidth]{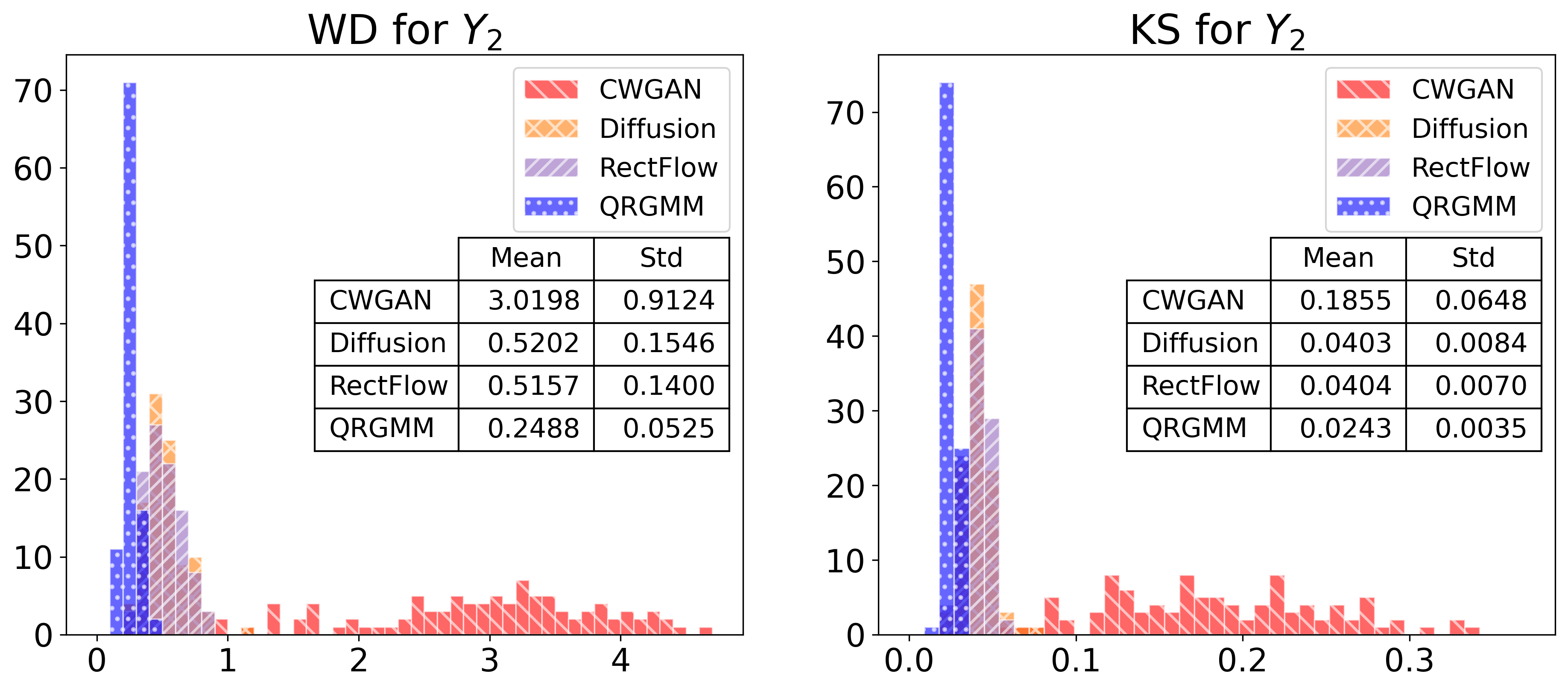}
		\end{array}$
	}
	{Generated Samples Approximating Distribution of $Y_1$ and $Y_2$: WD and KS.\label{fig_ECS_vs_hist_WDKS_Y12}}
	{}
\end{figure}

\begingroup
\setlength{\bibsep}{0pt plus 0.3ex}

\endgroup

\end{document}